\PassOptionsToPackage{table,dvipsnames}{xcolor}

\documentclass[10pt,twocolumn,letterpaper]{article}

\usepackage[pagenumbers]{iccv} 

\usepackage{graphicx} 
\usepackage{subcaption}   
\usepackage{array}    
\usepackage{booktabs} 
\usepackage{xcolor}   
\usepackage{float}    
\usepackage{enumitem}
\usepackage{pgfplots}
\pgfplotsset{compat=1.18}

\usepackage{multirow}

\newcommand{\ours}{\texttt{DIP}\xspace}

\definecolor{baselinecolor}{rgb}{0.9, 0.9, 1.0}
\newcommand{\baseline}[1]{\cellcolor{baselinecolor}{#1}}

\usepackage{pifont}%
\newcommand{\cmark}{\ding{51}}%
\newcommand{\xmark}{\ding{55}}%

\newcommand{\tablestyle}[2]{\setlength{\tabcolsep}{#1}\renewcommand{\arraystretch}{#2}\centering\footnotesize}

\newenvironment{customlegend}[1][]{%
    \begingroup
    \csname pgfplots@init@cleared@structures\endcsname
    \pgfplotsset{#1}%
}{%
    \csname pgfplots@createlegend\endcsname
    \endgroup
}%

\def\addlegendimage{\csname pgfplots@addlegendimage\endcsname}

\definecolor{col0}{RGB}{199,199,199}  
\definecolor{col1}{RGB}{255,127,14}   
\definecolor{col2}{RGB}{44,160,44}    
\definecolor{col3}{RGB}{214,39,40}    
\definecolor{col4}{RGB}{148,103,189}  
\definecolor{col5}{RGB}{140,86,75}    
\definecolor{col6}{RGB}{227,119,194}  
\definecolor{col7}{RGB}{127,127,127}  
\definecolor{col8}{RGB}{188,189,34}   
\definecolor{col9}{RGB}{23,190,207}   
\definecolor{col10}{RGB}{174,199,232} 
\definecolor{col11}{RGB}{255,187,120} 
\definecolor{col12}{RGB}{152,223,138} 
\definecolor{col13}{RGB}{255,152,150} 
\definecolor{col14}{RGB}{197,176,213} 
\definecolor{col15}{RGB}{196,156,148} 
\definecolor{col16}{RGB}{247,182,210} 
\definecolor{col17}{RGB}{31,119,180}  
\definecolor{col18}{RGB}{219,219,141} 
\definecolor{col19}{RGB}{158,218,229} 

\newcommand{\parag}[1]{\smallskip\noindent\textbf{#1}\enspace}

\definecolor{iccvblue}{rgb}{0.21,0.49,0.74}
\usepackage[pagebackref,breaklinks,colorlinks,allcolors=iccvblue]{hyperref}

\def\paperID{3880} 
\def\confName{ICCV}
\def\confYear{2025}

\title{DIP: Unsupervised \underline{D}ense \underline{I}n-Context \underline{P}ost-training of Visual Representations}

\author{%
Sophia Sirko-Galouchenko\textsuperscript{1,2} 
\hspace{2.5mm} 
Spyros Gidaris\textsuperscript{1}
\\
\vspace{3.5mm}
Antonin Vobecky\textsuperscript{1,3,4}
\hspace{2.5mm} 
Andrei Bursuc\textsuperscript{1}
\hspace{2.5mm} 
Nicolas Thome\textsuperscript{2,5}
\and
\textsuperscript{1}Valeo.ai \hspace{2.5mm} 
\textsuperscript{2}Sorbonne Universit\'e, CNRS, ISIR, F-75005 Paris, France \\
\textsuperscript{3}FEE CTU \hspace{2.5mm} 
\textsuperscript{4}CIIRC CTU Prague \hspace{2.5mm}
\textsuperscript{5} Institut universitaire de France (IUF)
}

\begin{document}
\makeatletter
\let\@oldmaketitle\@maketitle
\renewcommand{\@maketitle}{\@oldmaketitle
  \begin{center}
  \small
  \begin{tabular}{cccccc}
  \centering
    \includegraphics[width=0.13\textwidth]{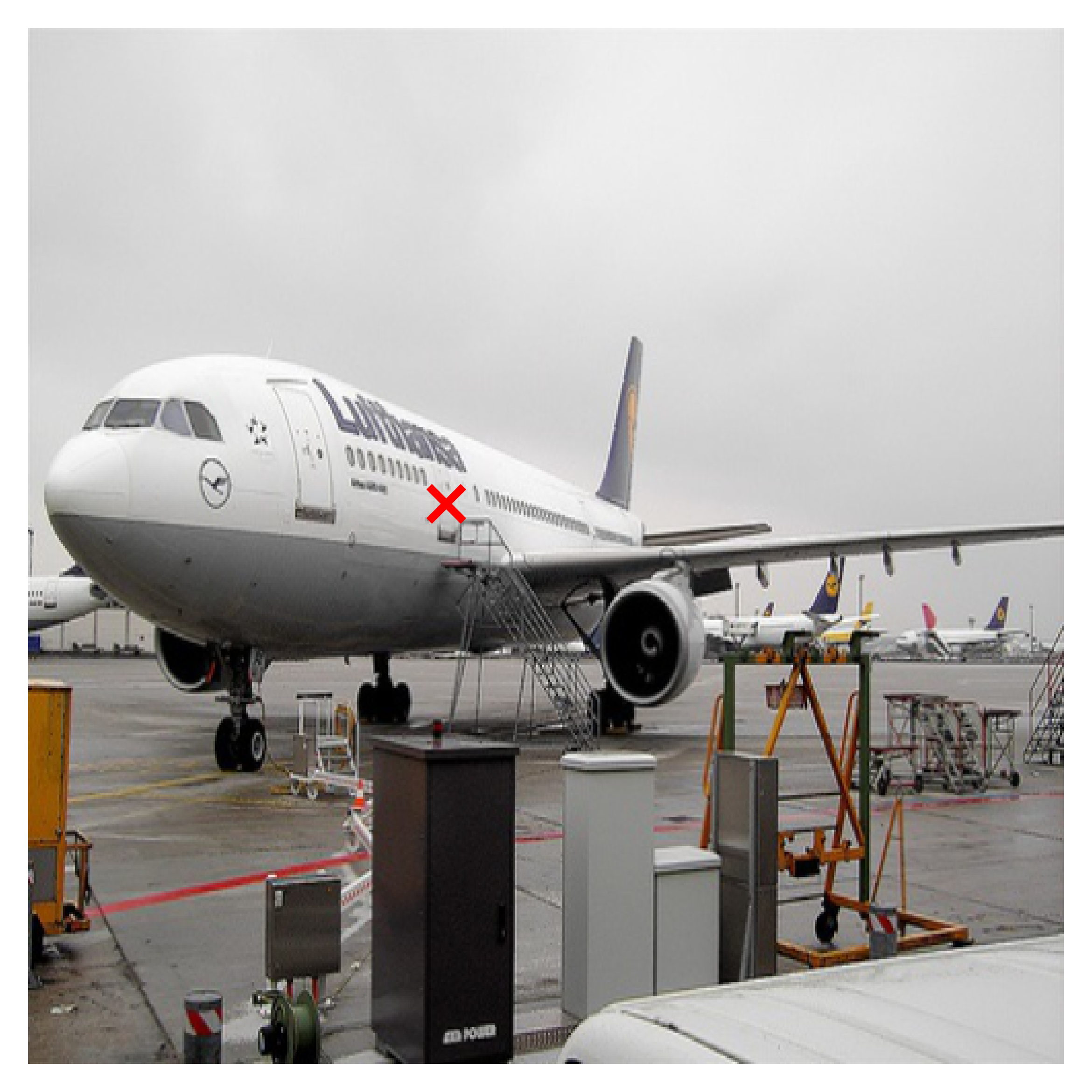} & 
    \includegraphics[width=0.13\textwidth]{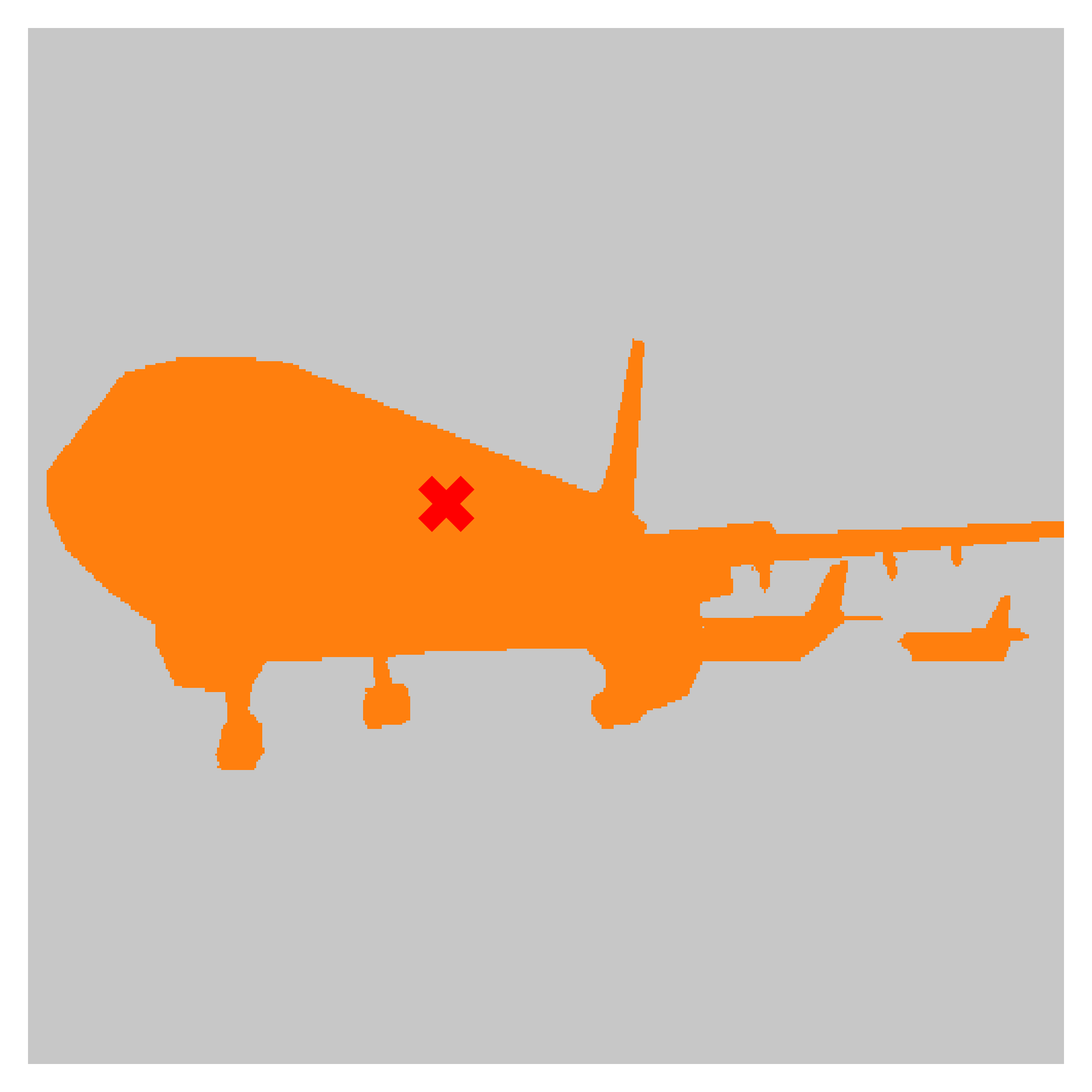} & 
    \includegraphics[width=0.13\textwidth]{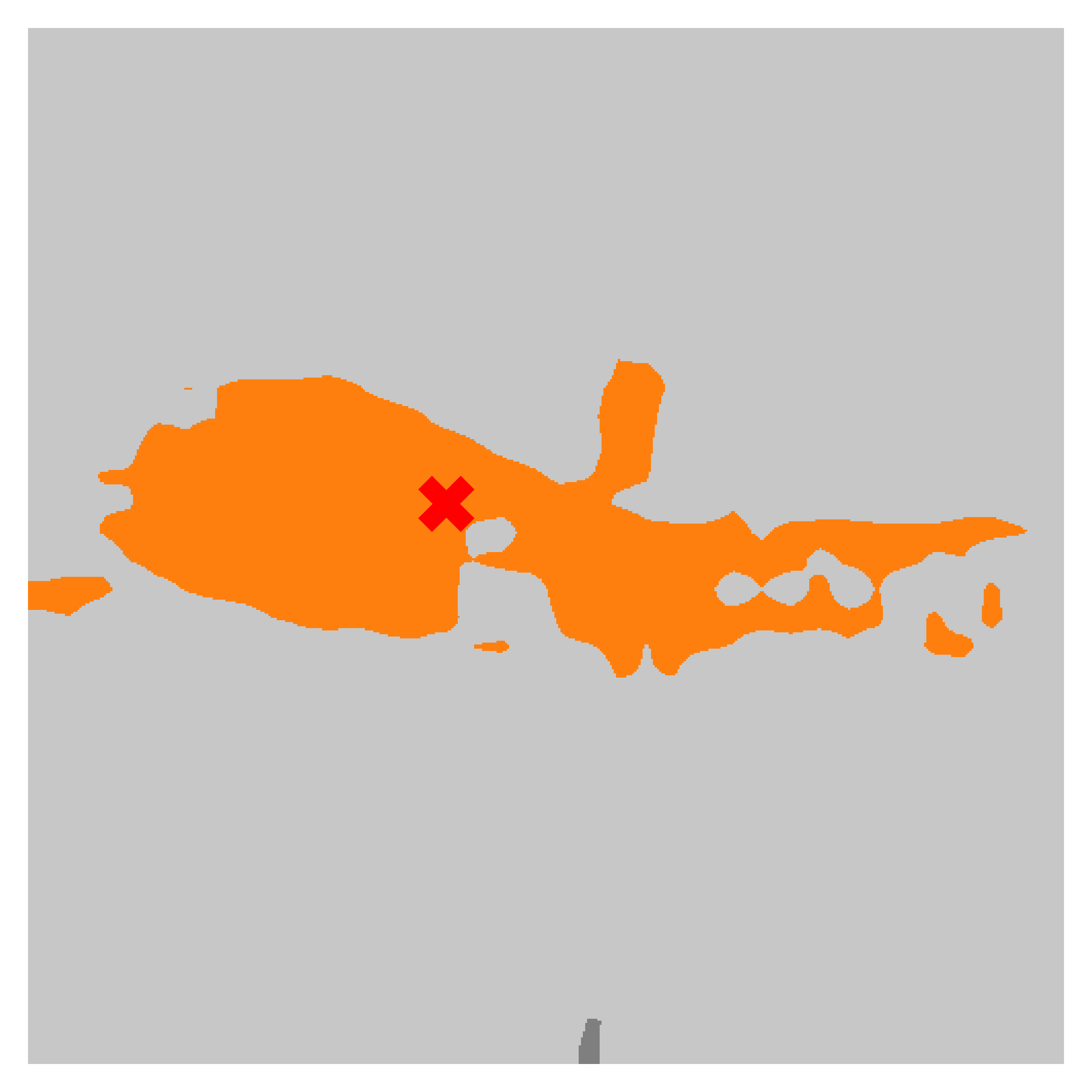} & 
    \includegraphics[width=0.13\textwidth]{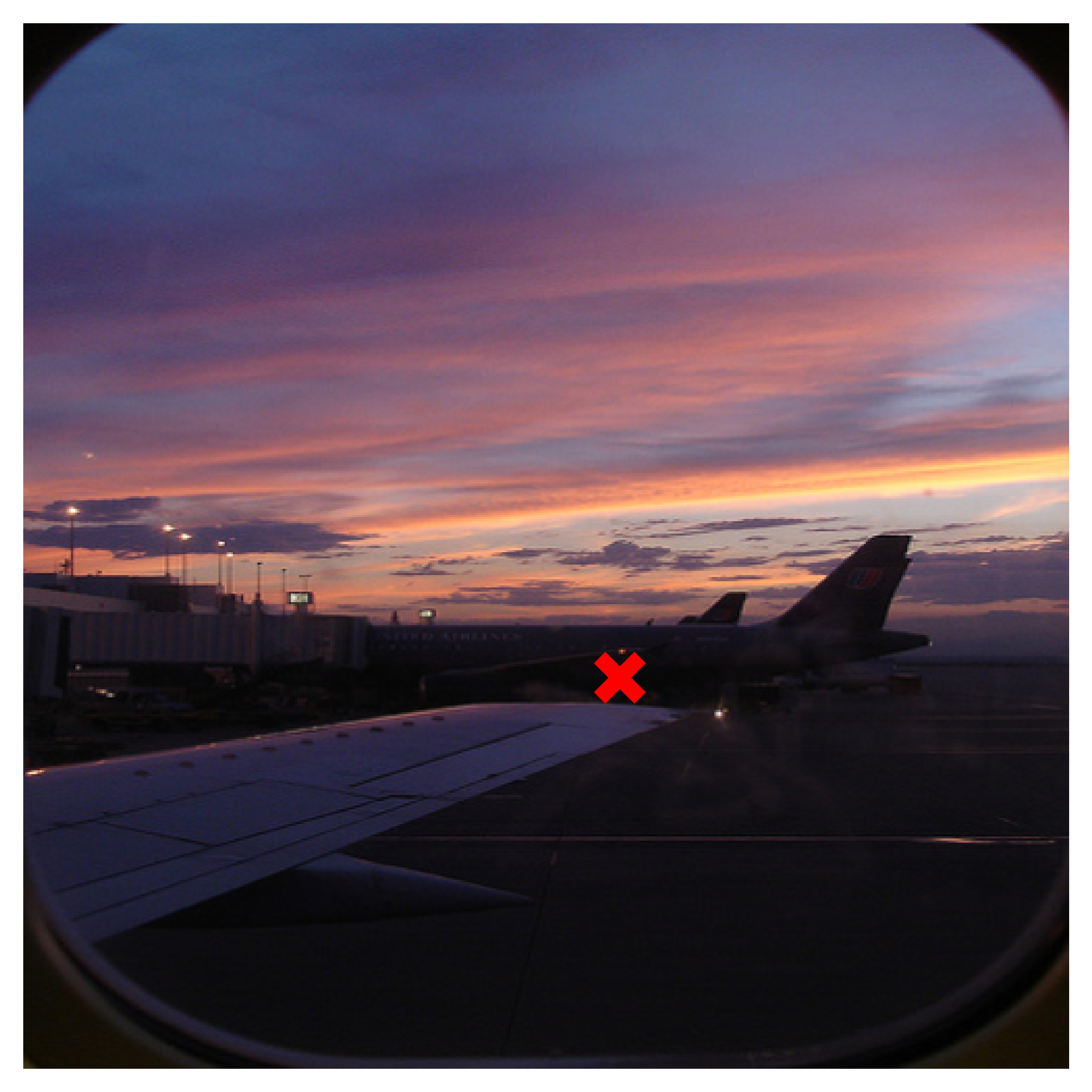} & 
    \includegraphics[width=0.13\textwidth]{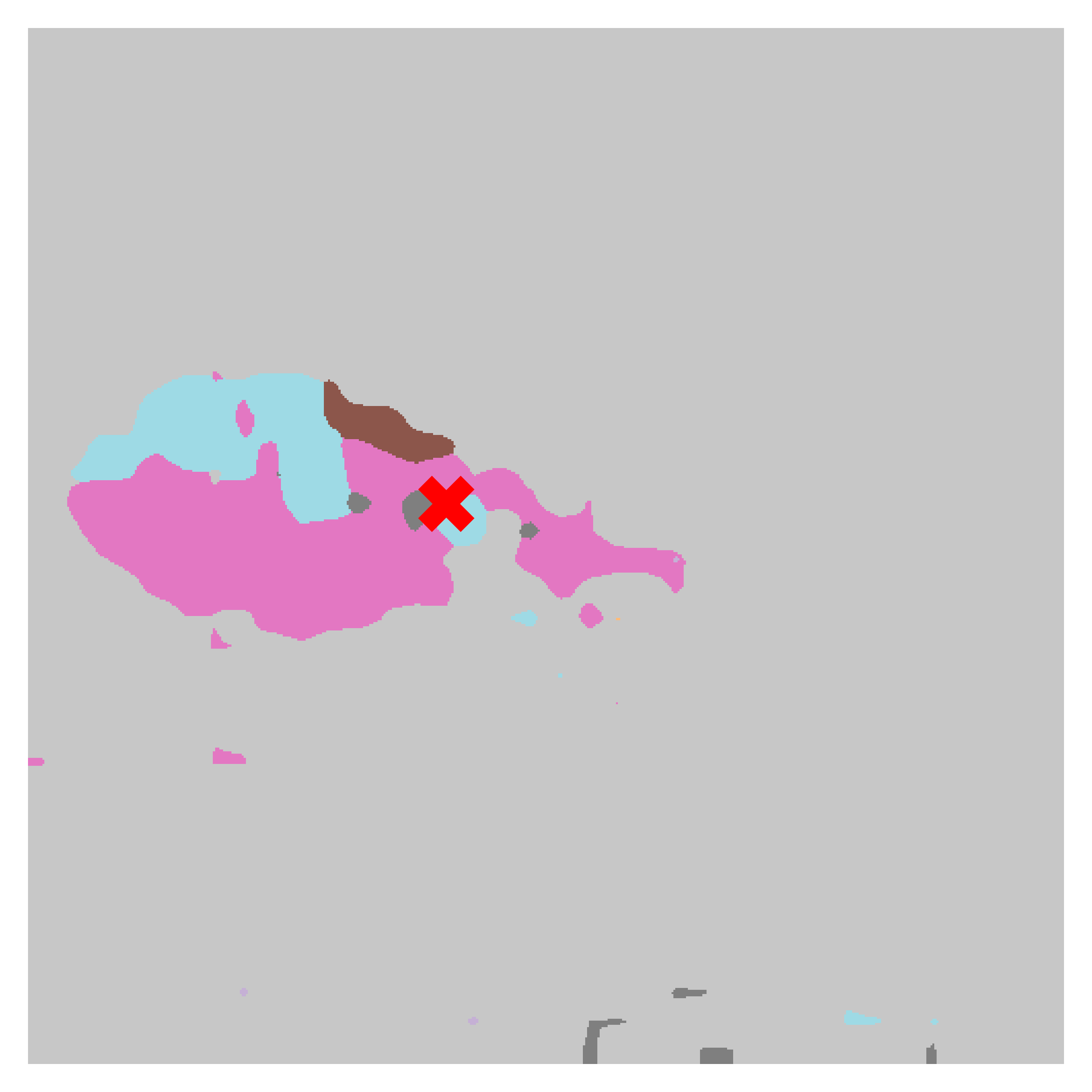} &
    \includegraphics[width=0.13\textwidth]{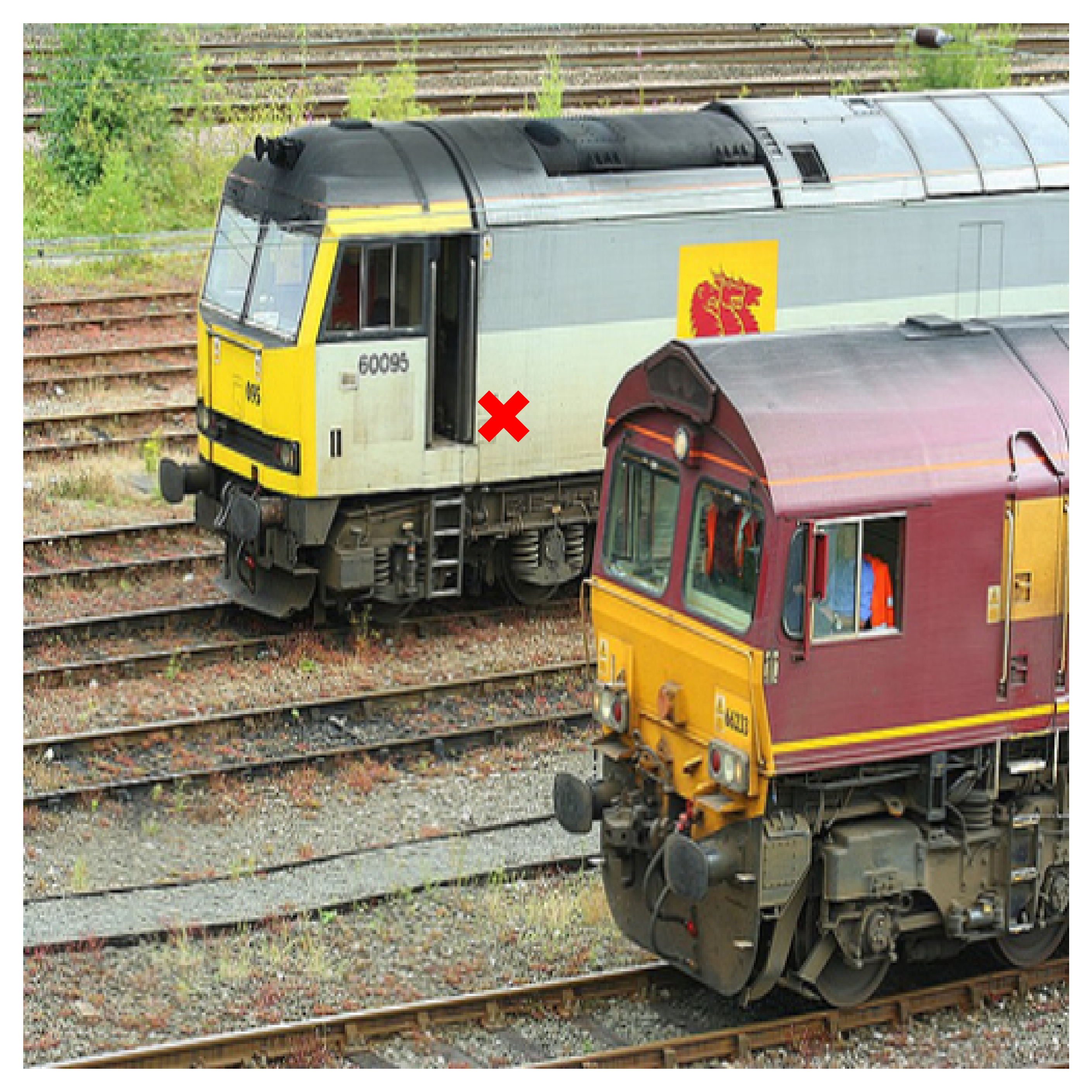} 
    \\ 
    \includegraphics[width=0.13\textwidth]{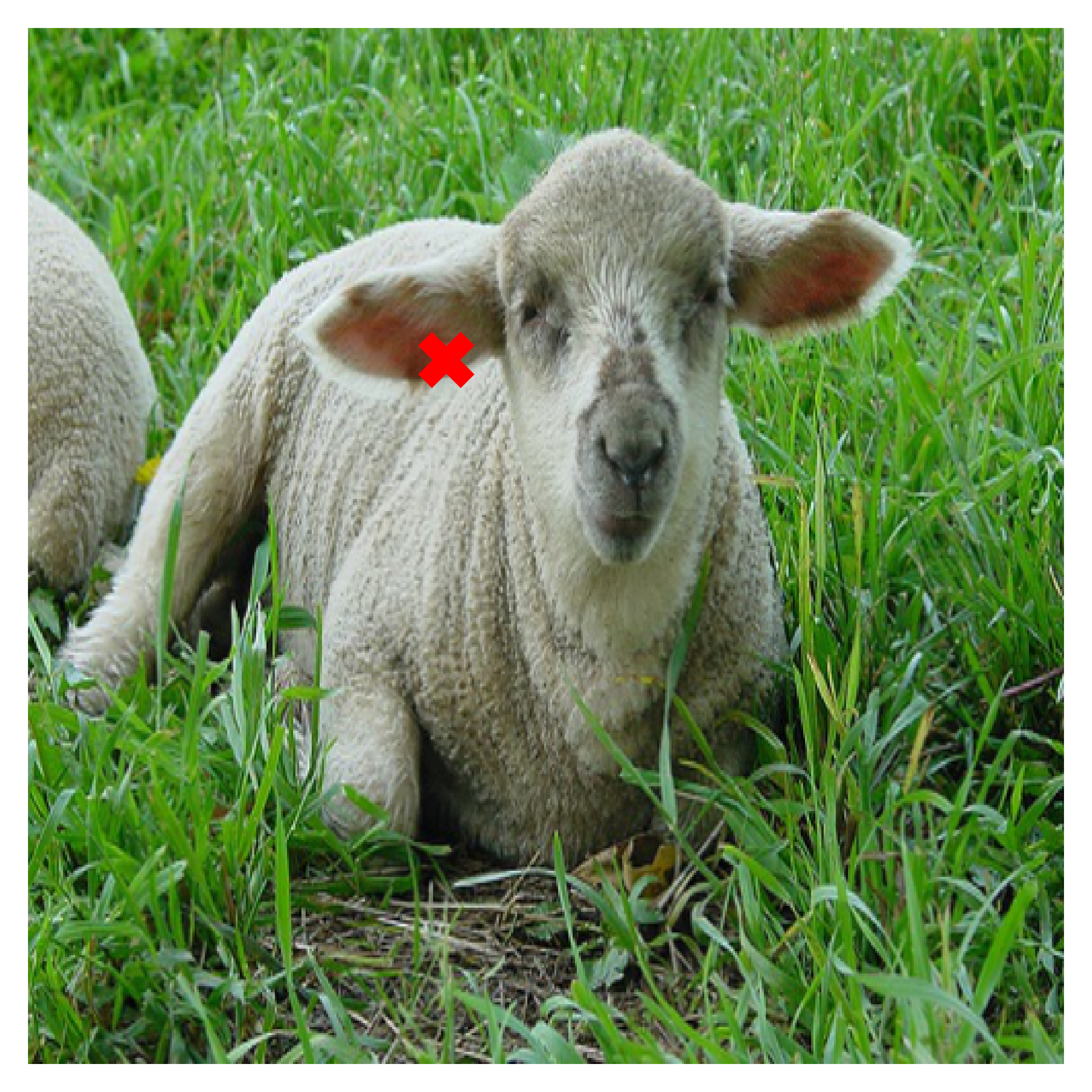} & 
    \includegraphics[width=0.13\textwidth]{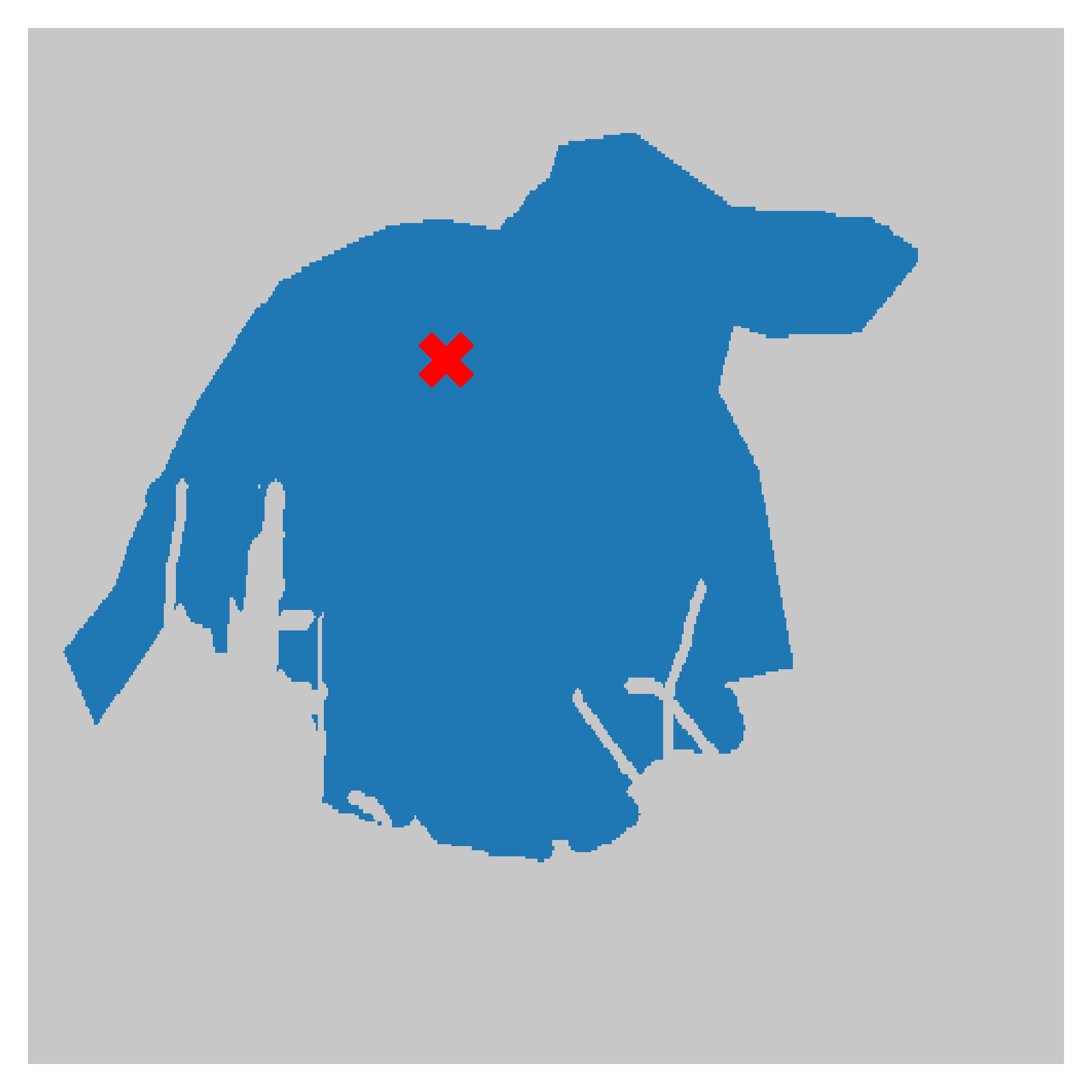} & 
    \includegraphics[width=0.13\textwidth]{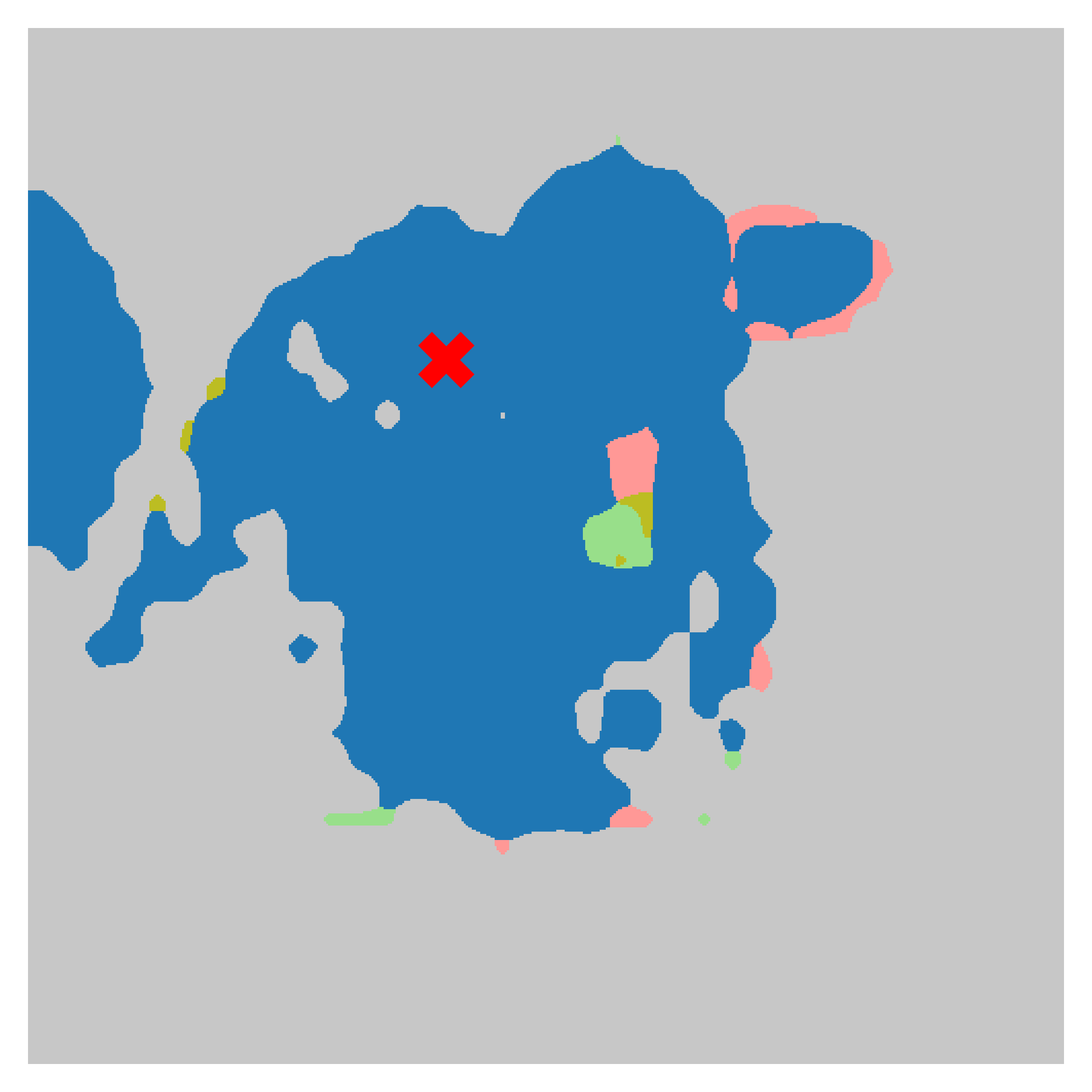} & 
    \includegraphics[width=0.13\textwidth]{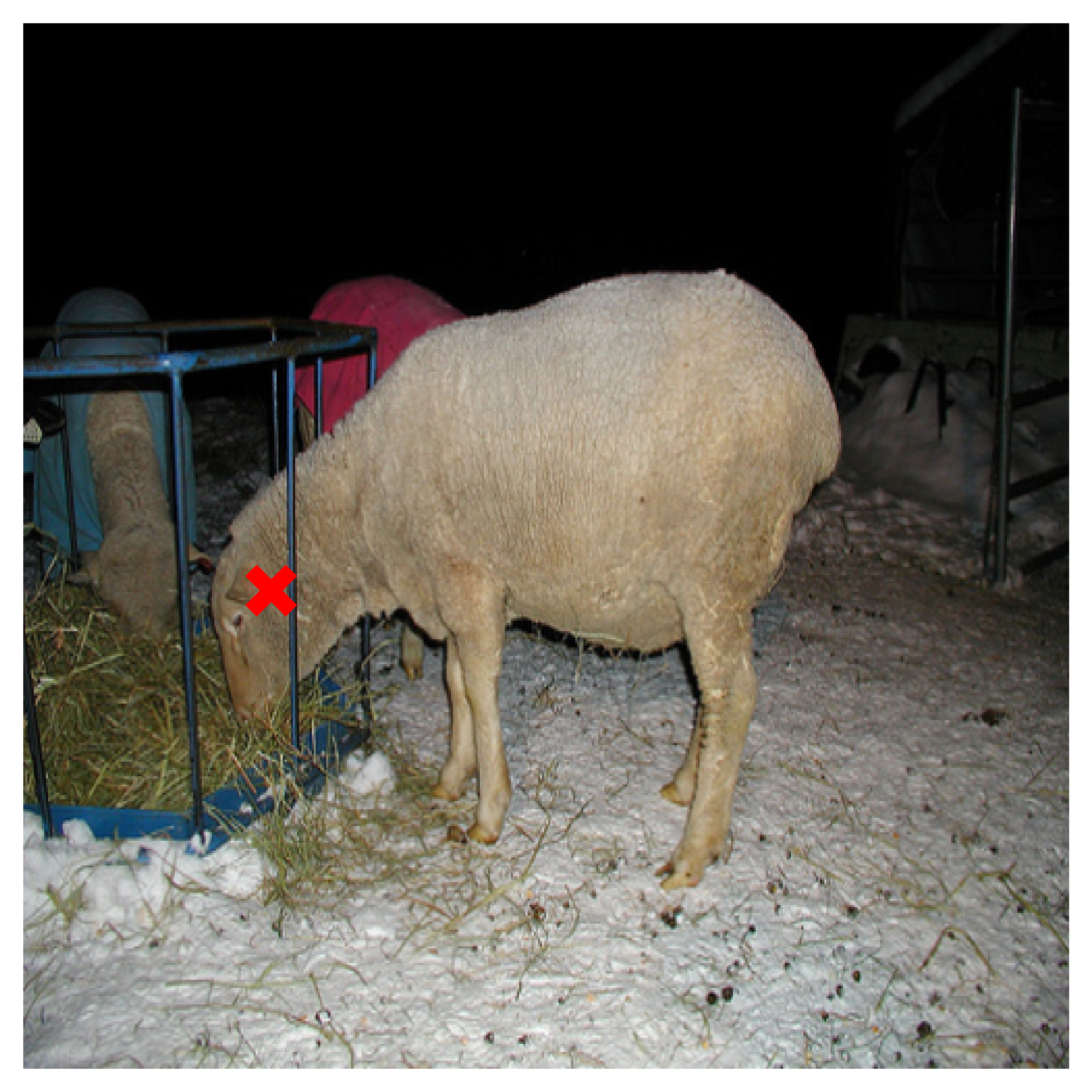} & 
    \includegraphics[width=0.13\textwidth]{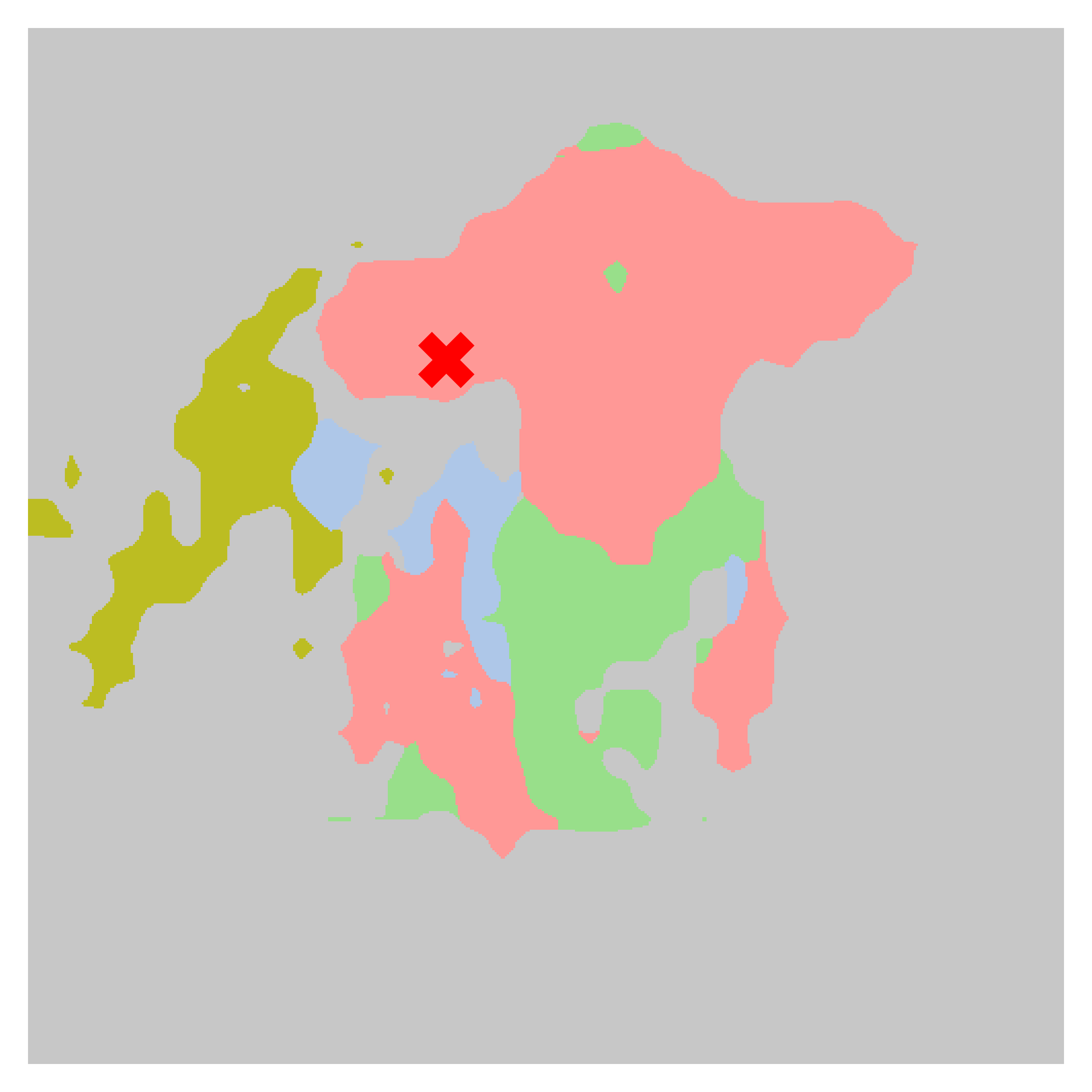} &
    \includegraphics[width=0.13\textwidth]{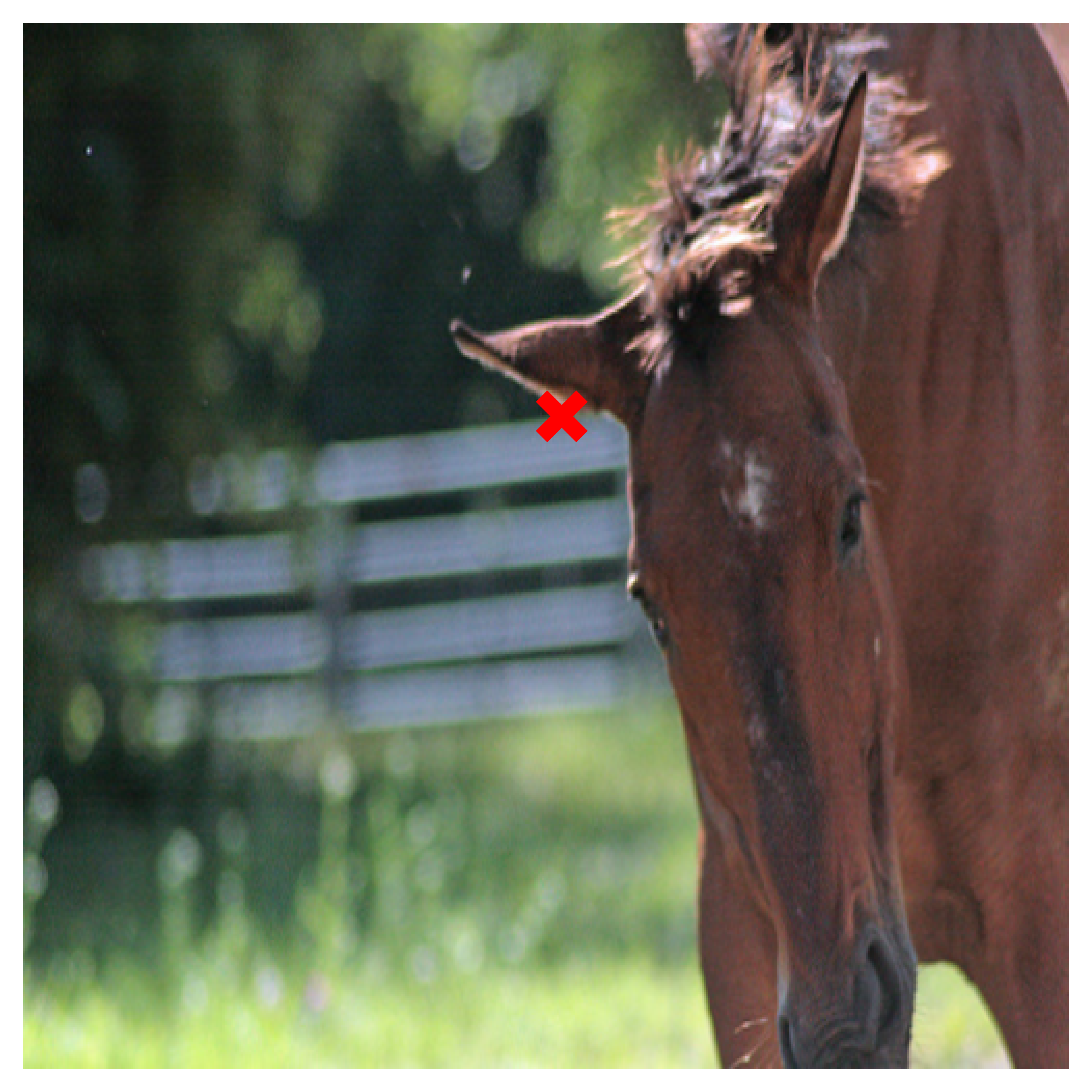}\\
    
    (a) Query & (b) Ground Truth & (c) \ours output & (d) \ours 1$^{\text{st}}$ NN & (e) Dv2R output & (f) Dv2R 1$^{\text{st}}$ NN\\
  \end{tabular}
  \vspace{0.03cm}
  \begin{tabular}{*{12}{c}}
    {\textcolor{col8}{\textbf{\normalsize cat}}} & 
    {\textcolor{col12}{\textbf{\normalsize dog}}} & 
    {\textcolor{col13}{\textbf{\normalsize horse}}} & 
    {\textcolor{col17}{\textbf{\normalsize sheep}}} & 
    {\textcolor{col10}{\textbf{\normalsize cow}}} & 
    {\textcolor{col1}{\textbf{\normalsize aeroplane}}} & 
    {\textcolor{col7}{\textbf{\normalsize car}}} & 
    {\textcolor{col6}{\textbf{\normalsize bus}}} & 
    {\textcolor{col5}{\textbf{\normalsize bottle}}} & 
    {\textcolor{col11}{\textbf{\normalsize diningtable}}} & 
    {\textcolor{col14}{\textbf{\normalsize motorbike}}} & 
    {\textcolor{col19}{\textbf{\normalsize train}}}
  \end{tabular}
  \vspace{-5pt}
  \captionof{figure}{\textbf{Dense retrieval-based semantic segmentation in low-shot regimes (40 examples) using patch feature similarities.} For each query image (a), we show the ground truth (b), our \ours model's prediction (c), and its retrieved neighbor (d) for the patch marked with a red cross in (a). In (e) and (f), we display DINOv2R's (Dv2R) output and its nearest neighbor, respectively. Our \ours representations retrieve more coherent neighbors, yielding more accurate segmentations than DINOv2R.}\label{fig:segmentation_comparison}
\end{center}}
\makeatother
\maketitle
\begin{abstract}
We introduce \ours, a novel unsupervised post-training method designed to enhance dense representations in large-scale pretrained vision encoders for in-context scene understanding. Unlike prior approaches using complex self-distillation architectures, our method trains the vision encoder using pseudo-tasks that simulate downstream in-context scenarios, inspired by meta-learning principles.
To enable post-training on unlabeled data, we propose an automatic mechanism for generating in-context tasks that combines a pretrained diffusion model and the vision encoder.
\ours is simple, unsupervised, and computationally efficient, requiring under 9 hours on a single A100 GPU. By learning dense representations through pseudo in-context tasks, it achieves strong performance across a variety of downstream real-world in-context scene understanding tasks. It outperforms both the initial vision encoder and prior methods, offering a practical and effective solution for improving dense representations. See code \href{https://github.com/sirkosophia/DIP}{here}
\end{abstract}
\vspace{-15pt}    
\section{Introduction}
\label{sec:intro}

Our goal is to learn dense image representations from unlabeled data for effective in-context scene understanding. In-context learning, which enables models to adapt to tasks without updating parameters, has seen remarkable success in large language models (LLMs) \cite{brown2020language}. By providing a few examples within the input prompt, LLMs generalize effectively to new tasks. Inspired by this, recent efforts aim to bring similar in-context learning capabilities to vision-only models \cite{hummingbird, bar2022visualpromptingimageinpainting}. A notable approach by Balazevic et al.~\cite{hummingbird} reformulates dense prediction tasks as nearest-neighbor retrieval problems using patch feature similarities, demonstrating strong performance, especially with limited data. However, while recent self-supervised Vision Transformers (ViTs)~\cite{hummingbird, dinov2, venkataramanan2025francanestedmatryoshkaclustering} show promise for in-context scene understanding, they still fall short of matching LLMs’ in-context learning success. To bridge this gap, it is crucial to learn dense features that establish strong semantic correspondences between patches in test and training images.

A key insight from prior work~\cite{neco, ziegler2022self, salehi2023time} is that leveraging pretrained Vision Foundation Models (VFMs), such as DINOv2R~\cite{dinov2,darcet2024vision}, and specializing them through unsupervised post-training is more computationally efficient and effective than training from scratch. This aligns with trends in LLMs, where large-scale pretraining is followed by cheaper post-training stages to specialize in specific skills of interest. We refer to the pretrained VFM undergoing post-training as the \emph{base model}.

Most methods, whether trained from scratch or post-trained, rely on self-distillation, where the teacher is an exponential moving average of student weights~\cite{grill2020bootstrap}. These frameworks use architectures and objectives designed to promote dense nearest-neighbor retrieval skills. 
For instance, state-of-the-art methods enforce object consistency across images and views~\cite{cribo} or between patch rankings from two random views of the same image~\cite{neco}.
However, self-distillation often suffers from poor stability \cite{zhou2022ibot} and hyperparameter sensitivity \cite{assran2022masked}. Moreover, these approaches introduce additional complexity, such as differentiable sorting \cite{neco}, RoI alignment units \cite{neco, ziegler2022self, he2017mask}, Sinkhorn-Knopp optimization \cite{ziegler2022self, salehi2023time}, or learnable patch pooling \cite{hummingbird, cribo}. 
While improving performance, these components make the methods harder to interpret and tune, and less transparent in terms of their underlying mechanisms.

In this work, we adopt a post-training approach due to its computational efficiency and effectiveness. Within this context, we address two key questions: (1) \emph{Are there simpler and more effective alternatives to existing self-distillation frameworks with complex architectures and objectives?}, and  (2)~\emph{Beyond the base model, can we leverage other vision foundation models to achieve our goal?} 

We propose a simpler post-training approach with an unsupervised 
learning objective explicitly designed for in-context scene understanding. Inspired by meta-learning, which learns general knowledge from diverse tasks to address unseen but related tasks, we automatically construct multiple in-context tasks from unlabeled data and train the model on them. This enables the model to acquire transferable dense representations for efficiently solving new in-context tasks during downstream stages. Unlike prior methods, our approach eliminates complex training objectives and architectures, and avoids self-distillation frameworks.

To achieve unsupervised post-training, we generate in-context pseudo-tasks that mimic the tasks encountered during downstream use. Each pseudo-task consists of support examples with segmentation masks and a query image segmented using the support examples. The support set includes a positive example (sharing objects with the query) and ``distractor" examples (likely unrelated). Our pseudo-tasks resemble meta-learning \emph{episodes}~\cite{vinyals2016matching} 
with the difference that we do not use any manual annotations.

Instead, we leverage pretrained VFMs to construct these pseudo-tasks. Specifically, apart from the base model (DINOv2R) that we post-train, we use an auxiliary pretrained generative model, Stable Diffusion (SD)~\cite{ssd1b}. 
SD provides high-fidelity, class-agnostic image segments in an unsupervised manner, while the base model retrieves candidate positives for each query and assigns pseudo-labels to the segments.
Although SD is pre-trained on image-caption pairs, we consider our post-training unsupervised as we do not use captions to extract image segments from SD, and neither SD nor the base model rely on dense segmentation labels.

The main contributions of this paper are as follows:
\begin{itemize}
    \item We introduce \ours, a novel unsupervised post-training method that uses retrieval-based in-context learning to improve dense image representations. Unlike prior work, \ours is based on meta-training principles and eliminates the need for complex self-distillation architectures. Furthermore, \ours is computationally efficient, requiring less than 9 hours on a single A100 GPU for post-training.

    \item To enable post-training on unlabeled data, we propose an automatic mechanism for generating in-context tasks. In addition to the base model, this approach leverages the Stable Diffusion model (through the unsupervised, training-free DiffCut~\cite{diffcut} technique) to produce high-quality pseudo-segmentations from unlabeled images. 

    \item Experiments show that the learned dense representations generalize effectively to a wide variety of downstream tasks. These include six semantic segmentation datasets and one monocular depth prediction dataset, even in low-shot scenarios. Furthermore, we extend our approach to other VFMs, such as CLIP~\cite{clip}, and show that it consistently enhances their dense representations.

    \item Compared to DINOv2R, \ours produces more semantically coherent neighbors, leading to more accurate retrieval-based scene understanding, particularly in low-shot settings (see \cref{fig:segmentation_comparison}). Additionally, \ours outperforms recent state-of-the-art post-training methods in most retrieval-based tasks while being simpler.
\end{itemize}
\section{Related Work}
\label{sec:formatting}

\medskip\noindent\textbf{Self-supervised learning.} 
Self-supervised methods learn rich representations from vast amounts of raw data via pretext tasks providing supervision signals from the data, later enabling fine-tuning on downstream tasks with minimal annotations. 
The main self-supervised approaches for images include:
contrastive objectives that distinguish similar and dissimilar views~\cite{oord2018representation,chen2020simple,he2020momentum}, clustering-based objectives~\cite{asano2020self,caron2020unsupervised,gidaris2020learning}, self-distillation objectives that match representations across augmentations~\cite{caron2021emerging,grill2020bootstrap,bardes2022vicreg,gidaris2021obow,chen2021exploring}, and reconstruction objectives that predict pixel colors~\cite{he2022masked,xie2022simmim} or masked patch features from a teacher network~\cite{zhou2022ibot,assran2022masked, kakogeorgiou2022hide,gidaris2024moca,oquab2023dinov2,venkataramanan2025francanestedmatryoshkaclustering}. Recent approaches use pretrained VFMs for distillation in cooperation with self-supervised objectives~\cite{bao2022beit, dong2023peco, peng2022beit, hou2022milan}. 
These methods focus on learning global image representations for classification tasks. 
However, without full-finetuning or task decoders, they are less suitable for dense scene-understanding tasks like segmentation.
We build \ours on contrastive objectives~\cite{oord2018representation} equipped with a memory bank~\cite{he2020momentum}. 

\medskip\noindent\textbf{Dense self-supervision.} 
Learning localized image representations that can be rapidly adapted to scene understanding requires specialized self-supervised objectives that mimic downstream tasks
\cite{xie2021propagate, wang2021dense, henaff2021efficient, stegmuller2023croc, cribo, loca, hummingbird, henaff2022object, simoncini2024no}. One approach contrasts features from dissimilar pixels or pseudo-segments within images, improving object detection~\cite{xie2021propagate, wang2021dense, henaff2021efficient, henaff2022object, o2020unsupervised}. Others employ clustering strategies to produce object-aware supervision signals to improve semantic segmentation~\cite{stegmuller2023croc, ziegler2022self, loca}, while some extend patch-level contrastive learning across images~\cite{hummingbird} with nearest-neighbor consistency~\cite{cribo, stego, neco}.
Although we share similar objectives with \cite{neco,hummingbird} for in-context scene understanding, we introduce a novel self-supervised objective inspired by few-shot learning that leverages automatically computed object segment pseudo-labels.

Recent works build upon pretrained VFMs~\cite{ziegler2022self, stego, neco, sirko2024occfeat, karypidis2024dino} casting their self-supervised training as a post-training stage for preparing the VFM for dense downstream tasks. Similarly, \ours aims to rapidly endow pretrained VFMs with in-context dense reasoning skills.

\medskip\noindent\textbf{Unsupervised semantic segmentation.}
These works aim to produce object segments via self-supervised training. Earlier works employed clustering objectives, with segmentations refined via consistency across augmented views~\cite{kanezaki2018unsupervised, ji2019invariant,hwang2019segsort,picie,vobecky2025unsupervised}. Other works explore pretrained features from self-supervised VFMs \cite{stego,seong2023leveraging,lan2024smooseg,acseg,vobecky2025unsupervised} or diffusion UNet encoders \cite{tian2024diffseg,diffcut}. The recent DiffCut~\cite{diffcut} leverages recursive Normalized Cuts in the final attention layer of a pretrained diffusion UNet encoder to produce object segments without supervision. 
We use DiffCut to extract pseudo-segmentation maps for our in-context post-training, as these segments - while sometimes noisy - serve as effective proxies for actual object segments in our framework.

\medskip\noindent\textbf{Few-shot learning.} 
Few-shot learning develops models capable of rapid adaptation from limited labeled examples.
These models are typically trained on few-shot tasks (or \emph{episodes}) sampled from the training data, that simulate test-time conditions.
Existing approaches include memory-based methods that match queries to labeled examples~\cite{vinyals2016matching,mishra2018simple,santoro2016meta}, metric-learning techniques that learn distance functions from few samples~\cite{snell2017prototypical,sung2018learning}, optimization-based methods that learn efficient adaptation through few gradient steps~\cite{finn2017model, ravi2017optimization}, and generative approaches that predict classifier weights~\cite{qiao2018few, gidaris2019generating, gidaris2018dynamic}.
Inspired by Vinyals et al.~\cite{vinyals2016matching}, we devise a self-supervised pretext objective consisting of multiple in-context scene understanding tasks, each with a set of support samples stored in a memory and a query image to be segmented based on the support samples. In contrast to few-shot learning, we use no human annotation during post-training, generating labels automatically.

\section{Method}
\begin{figure*}[t]
    \centering

    \includegraphics[width=0.87\linewidth]{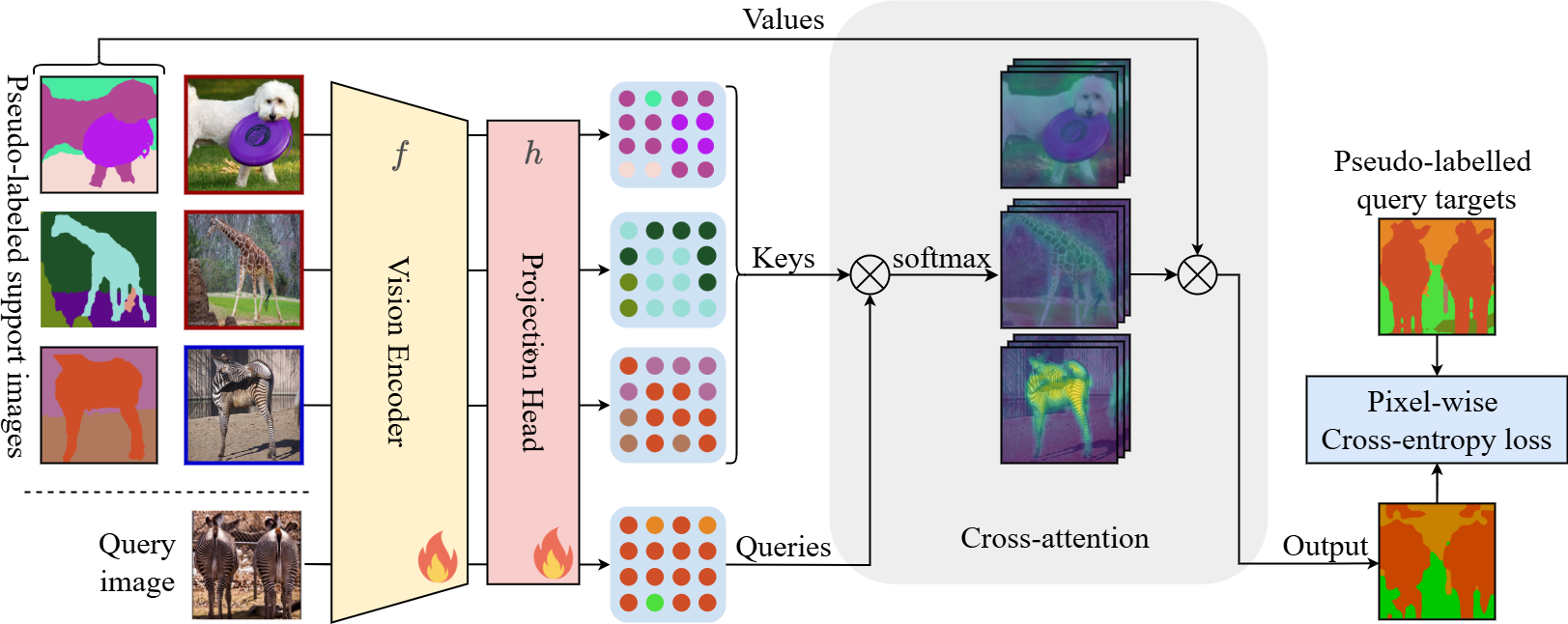}
    
    \vspace{-8pt}
    \caption{\textbf{Our unsupervised Dense In-context Post-training (\ours) method}.  
    During post-training, the model is given a pseudo in-context task, created automatically without human input. Each task includes a query image and a pseudo-labeled support set with a positive example (sharing objects with the query, as shown in the zebra example) and ``distractor" examples that contain different object categories. The model predicts the pseudo-labeled semantic segmentation of the query image using the support set as reference. To do this, it (1) extracts patch-wise features from both the query and support images using the vision encoder $f(\cdot)$ and projection head $h(\cdot)$; (2) computes segmentation predictions for the query image through cross-attention (query patch features as queries, support patch features as keys, and pseudo-labeled support patches as values). A pixel-wise cross-entropy loss is applied to the predictions. By training on these pseudo tasks, \ours enables the encoder to learn transferable dense representations, which are later used to efficiently solve new real in-context tasks.}
    \label{fig:pretraining method}
    \vspace{-5pt}
\end{figure*}

Our goal is to fine-tune a 
pretrained vision encoder, 
typically a ViT \cite{dosovitskiy2020image},
to produce dense features suitable for in-context dense prediction tasks framed as nearest-neighbor retrieval. Given a set of support images (i.e., training images) with semantic segmentation annotations and a query image (i.e., test image), we aim to ensure that patch-wise features extracted from the query image are highly similar (e.g., high cosine similarity) to support patches of the same object category, while being dissimilar to features of different objects.

To achieve this, we propose an unsupervised post-training approach (\cref{fig:pretraining method}) that fine-tunes the encoder on dense nearest-neighbor retrieval tasks automatically generated from unlabeled data. Training on pseudo-tasks that simulate real scenarios, the encoder learns robust features transferable to real in-context tasks.

In the following sections, we first detail our dense in-context post-training approach (\cref{sec:pretraining_objective}) and then describe the automatic process for constructing dense in-context tasks (\cref{sec:task_construction}), including the pseudo-labeling of class-agnostic segments and the selection of support examples.

\subsection{Dense In-Context Training}
\label{sec:pretraining_objective}

During training, we sample an in-context task $\mathcal{T}$ from dataset $\mathcal{D}$. Each task $\mathcal{T}$ consists of $K$ support examples $\mathcal{S} = \{(X_{s_i}, Y_{s_i})\}_{i=1}^{K}$ and one query example $(X_q, Y_q)$, where $X_q$ and $X_{s_i}$ are images, and $Y_q$ and $Y_{s_i}$ are their one-hot semantic segmentation pseudo-labels. Among the $K$ support examples, one is a ``positive" example sharing pseudo-classes with the query, while the remaining $K-1$ are ``distractor" examples unlikely to share any pseudo-class. Since our method is unsupervised, these tasks and pseudo-labels are automatically constructed (see \cref{sec:task_construction}).

\parag{Feature extraction.} Let $f(\cdot)$ be a pretrained encoder that our method fine-tunes. Given an input image $X$ (either query or support), $f(\cdot)$ produces patch-wise features $f(X) \in \mathbb{R}^{L \times D}$, where $L$ is the number of patches and $D$ is the feature dimension. We further apply a patch-wise multi-layer perceptron (MLP) network $h(\cdot)$ to project the features into a $D'$-dimensional space: $F = h(f(X)) \in \mathbb{R}^{L \times D'}$. The MLP includes an $\ell_{2}$-normalization layer at the end. Let $\theta$ denote the parameters of $(h \circ f)(\cdot)$ model.

\parag{Label pre-processing.} To align label resolution with patch features, we \emph{patchify} the one-hot pseudo-labels $Y_{s_i}$. We divide $Y_{s_i}$ into patches matching the encoder's patch size and average the one-hot labels within each patch, yielding $Y'_{s_i} \in \mathbb{R}^{L \times C}$, where $C$ is the number of pseudo-classes.

\parag{In-context dense predictions.} Using patch-wise features and patchified labels from the support set $\mathcal{S}$, we define a soft nearest-neighbor classification function $c_{\mathcal{S}}(\cdot)$ to predict dense labels $\hat{Y}_q = c_{\mathcal{S}}(X_q)$ for the query image $X_q$. First, we compute attention scores for each query patch over all $K \times L$ support patches:
\begin{equation} \label{eq:attention}
A = \text{softmax}\left(\frac{F_q \cdot F_S^T}{\tau}\right)\text{,}
\end{equation}
where $F_S \in \mathbb{R}^{(L \cdot K) \times D'}$ concatenates patch-wise features $F_{s_i}$ from all support images, and $\tau$ is the softmax temperature. The softmax normalization is over the support patches. The attention weights $A \in [0, 1]^{L \times (L \cdot K)}$ are used to compute a weighted average of the patchified support labels:
\begin{equation} \label{eq:waveraging}
\hat{Y}'_q = A \cdot Y'_S\text{,}
\end{equation}
where $Y'_S \in \mathbb{R}^{(L \cdot K) \times C}$ concatenates patchified labels $Y'_{s_i}$  from all support images, and $\hat{Y}'_q \in \mathbb{R}^{L \times C}$ represents the predicted patchified labels for query. 
Essentially, this defines a cross-attention layer \cite{vaswani2017attention}, with $F_q$ as queries, $F_S$ as keys, and $Y'_S$ as values.

Finally, we use nearest-neighbor interpolation to upsample $\hat{Y}'_q$ to the original image size, yielding the final label prediction $\hat{Y}_q = c_{\mathcal{S}}(X_q)$.

\parag{Training objective.} 
For a single task $\mathcal{T} {=} \{\mathcal{S},(X_q, Y_q)\}$, we minimize the pixel-wise cross-entropy loss $\mathcal{L}_{CE}(Y_q, c_{\mathcal{S}}(X_q), \theta)$ between the predicted labels $c_{\mathcal{S}}(X_q)$ and the pseudo-labels $Y_q$.
The model, comprising the pretrained encoder and randomly initialized MLP $(h \circ f)(\cdot)$ with parameters $\theta$, is trained by optimizing the expected loss over a collection of tasks sampled from $\mathcal{D}$:
\begin{equation} \label{sec:total_loss2}
\min_{\theta} \mathbb{E}_{\{\mathcal{S},(X_q, Y_q)\} \sim \mathcal{D}} \left[ \mathcal{L}_{CE}(Y_q, c_{\mathcal{S}}(X_q), \theta) \right]\text{.}
\end{equation}

\parag{Downstream stage.} 
For downstream in-context scene understanding tasks, we remove the head \( h(\cdot) \) and use the fine-tuned encoder \( f(\cdot) \). 
Following \cite{hummingbird,neco, pariza2024hbird}, we construct a larger memory bank (support set) than used during post-training, with 10,240,000 patch-wise features randomly sampled from the available training images for each dataset.  
For a query image, we extract its patch-wise features using \( f(\cdot) \), retrieve the top-30 nearest neighbors from the memory bank via cosine similarity, and apply cross-attention (as in \cref{eq:attention,eq:waveraging}) with temperature $\tau = 0.07$.

\begin{figure}[!t]
  \vspace{-8pt}
  \centering
  \begin{tabular}{@{}c@{\hspace{1pt}}c@{\hspace{1pt}}c@{\hspace{1pt}}c@{}}
    \includegraphics[width=0.11\textwidth]{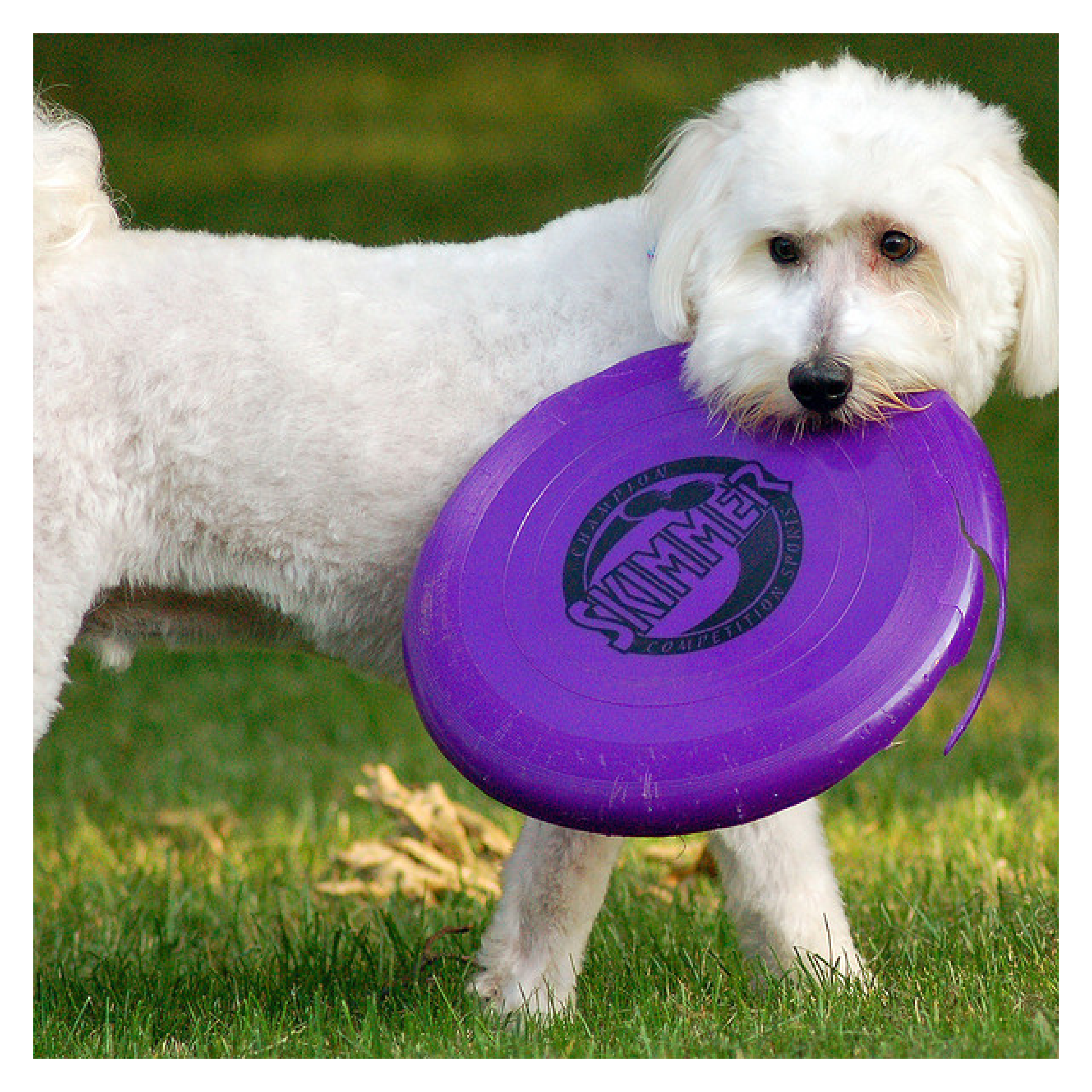} &
    \includegraphics[width=0.11\textwidth]{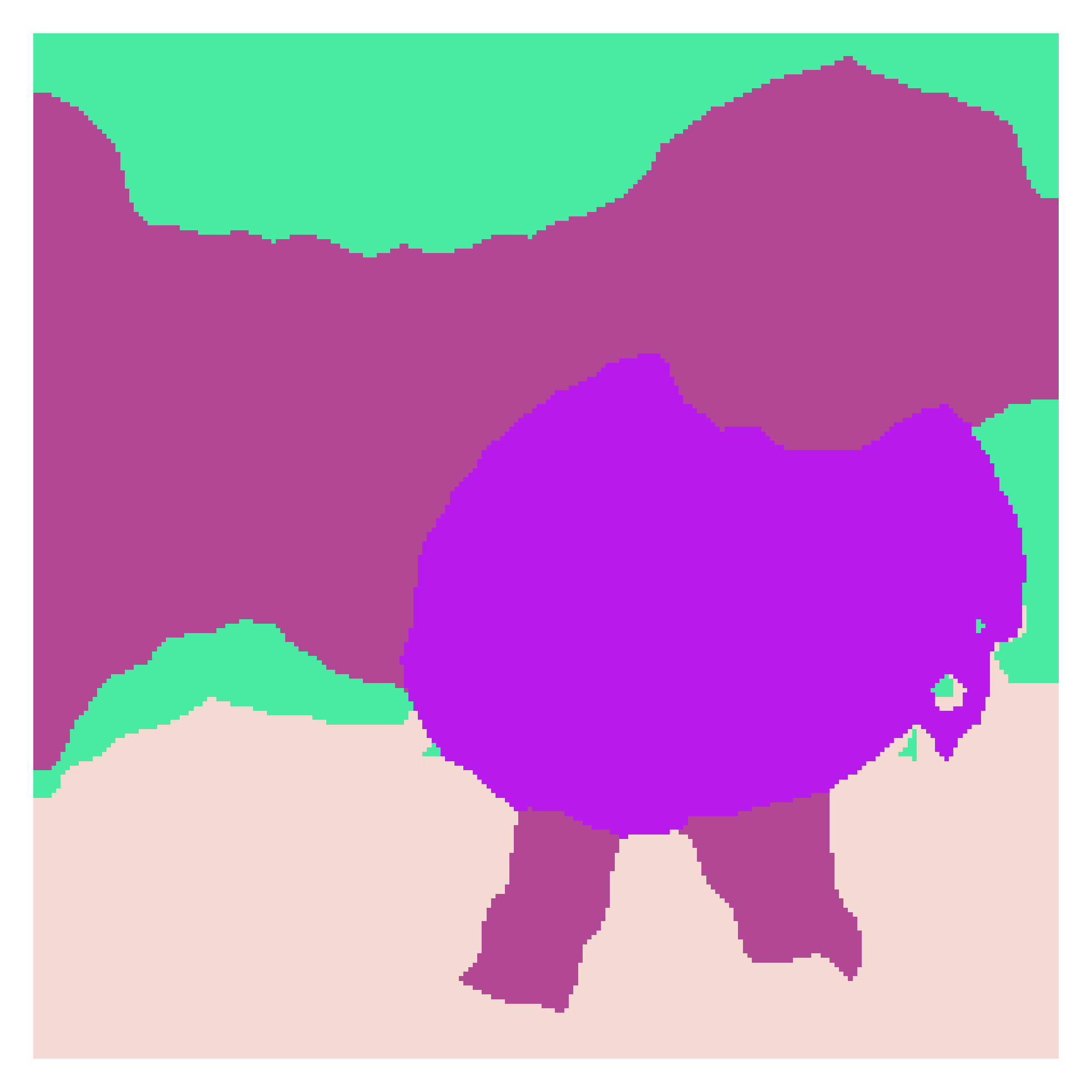} &
    \includegraphics[width=0.11\textwidth]{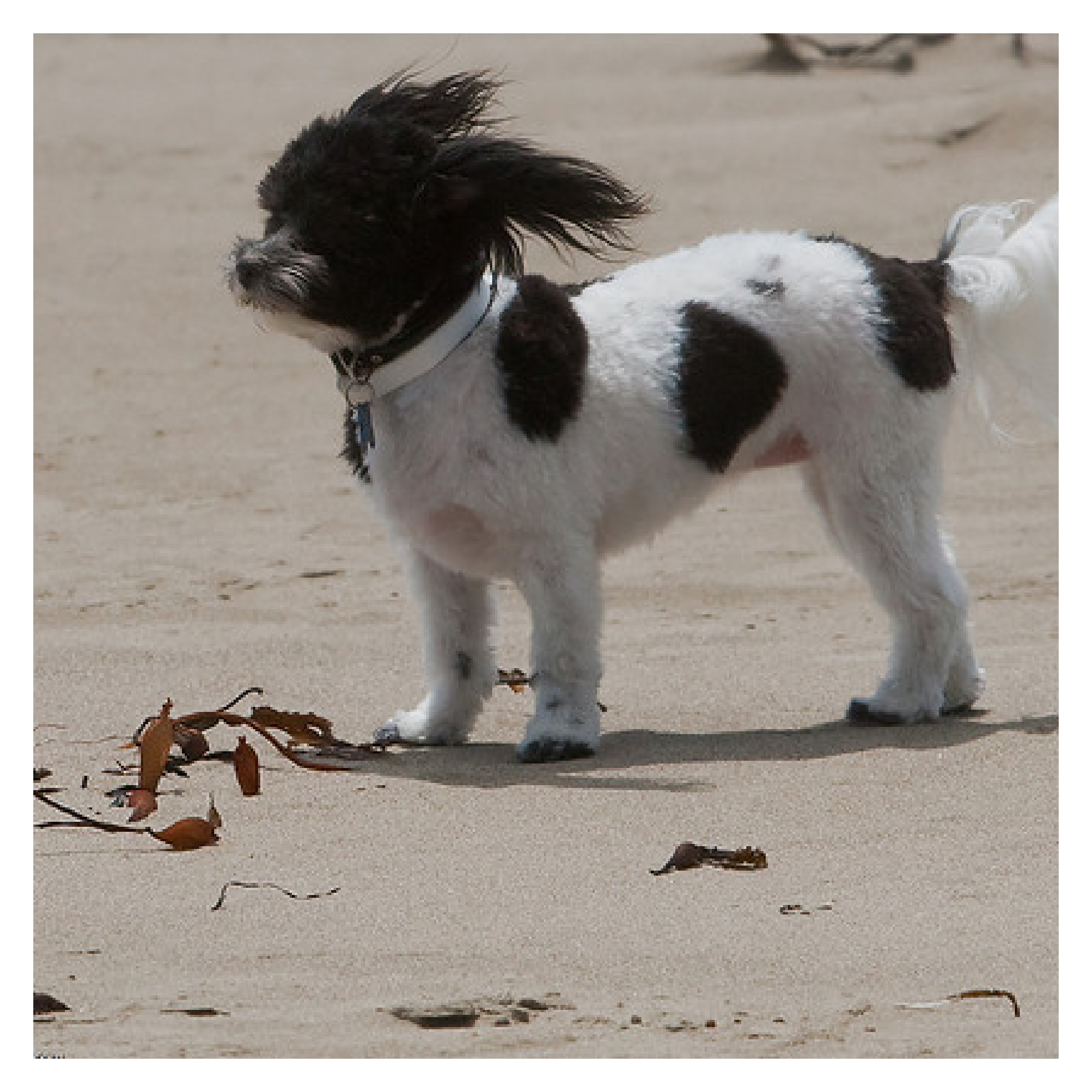} &
    \includegraphics[width=0.11\textwidth]{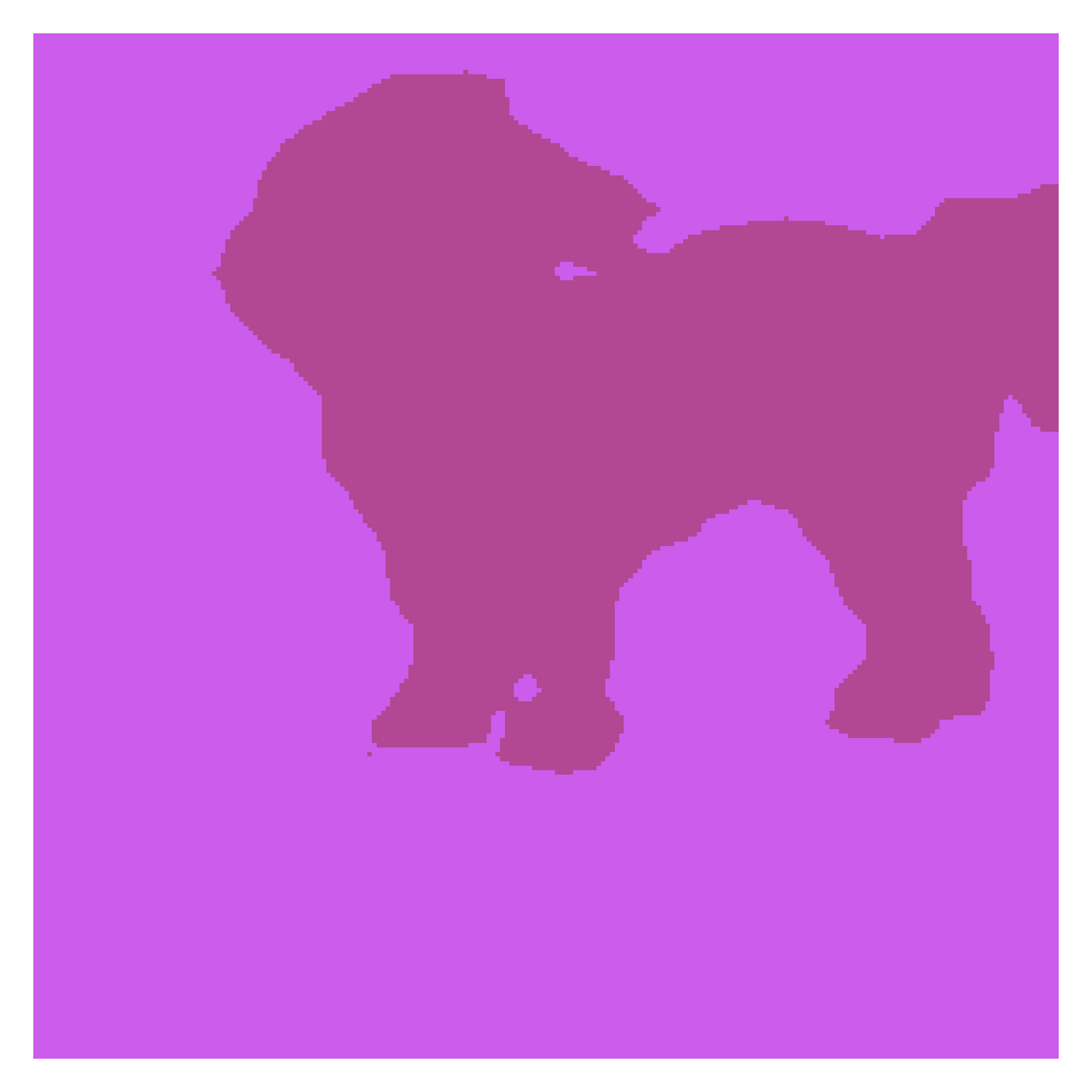} \\[-2pt]
    \includegraphics[width=0.11\textwidth]{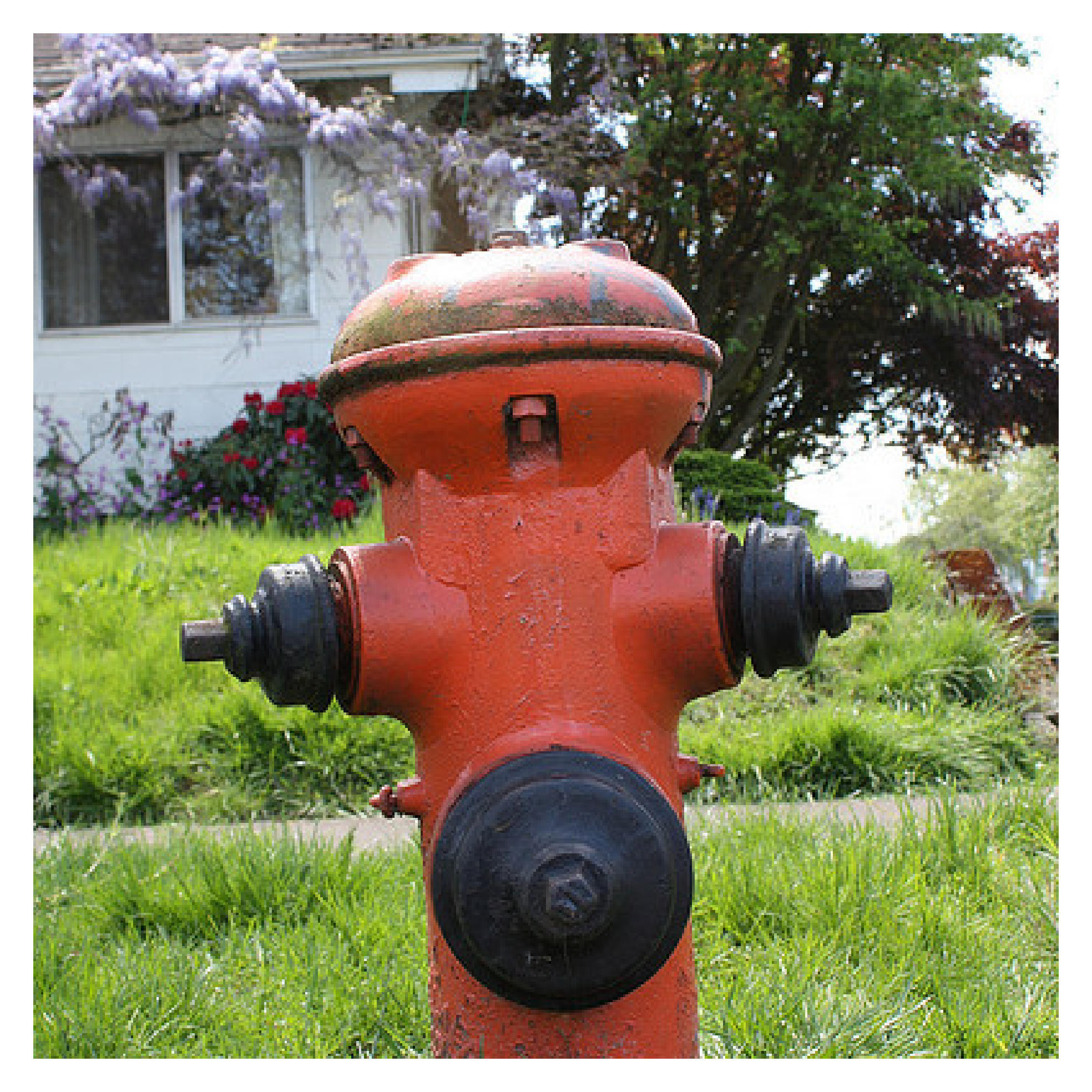} &
    \includegraphics[width=0.11\textwidth]{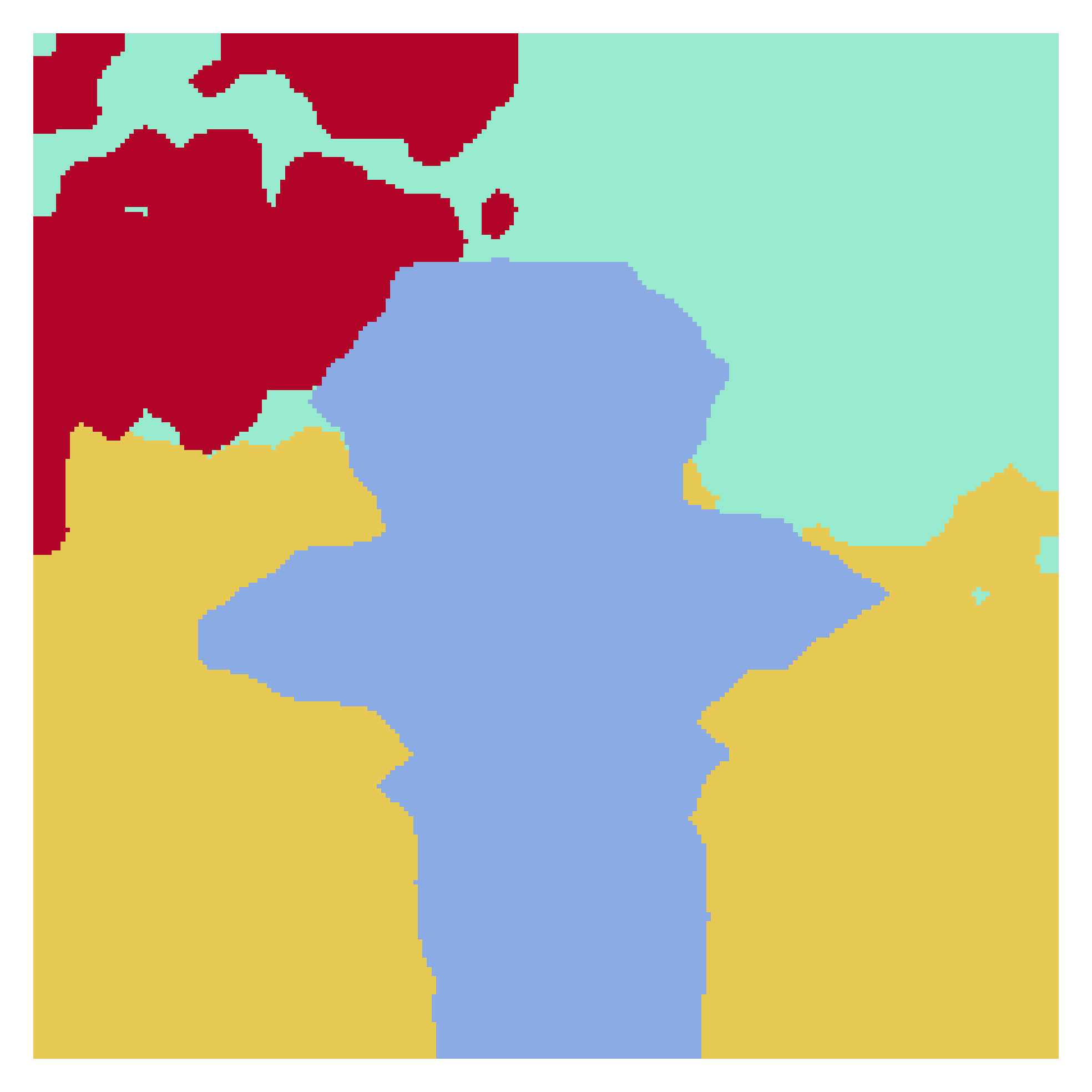} &
    \includegraphics[width=0.11\textwidth]{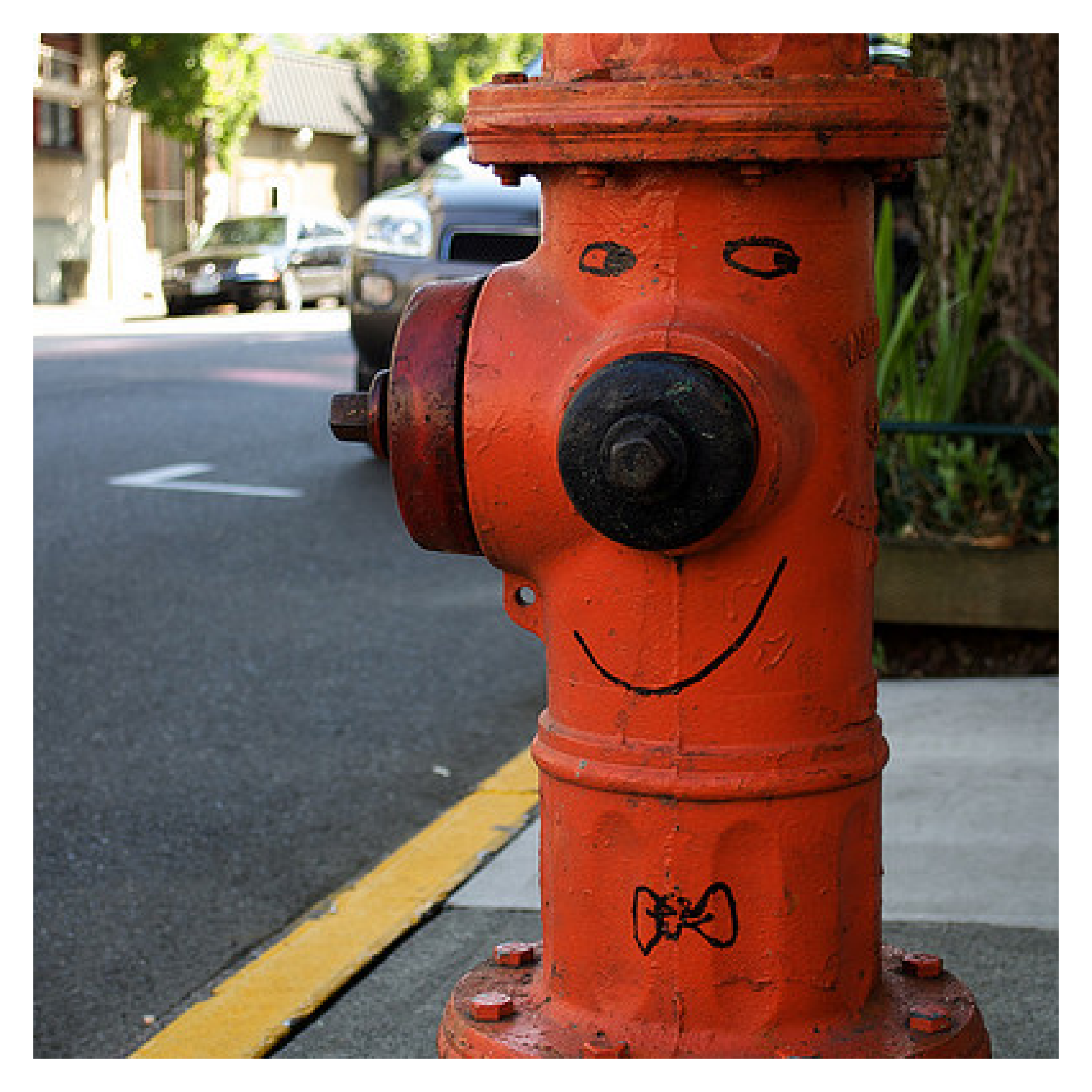} &
    \includegraphics[width=0.11\textwidth]{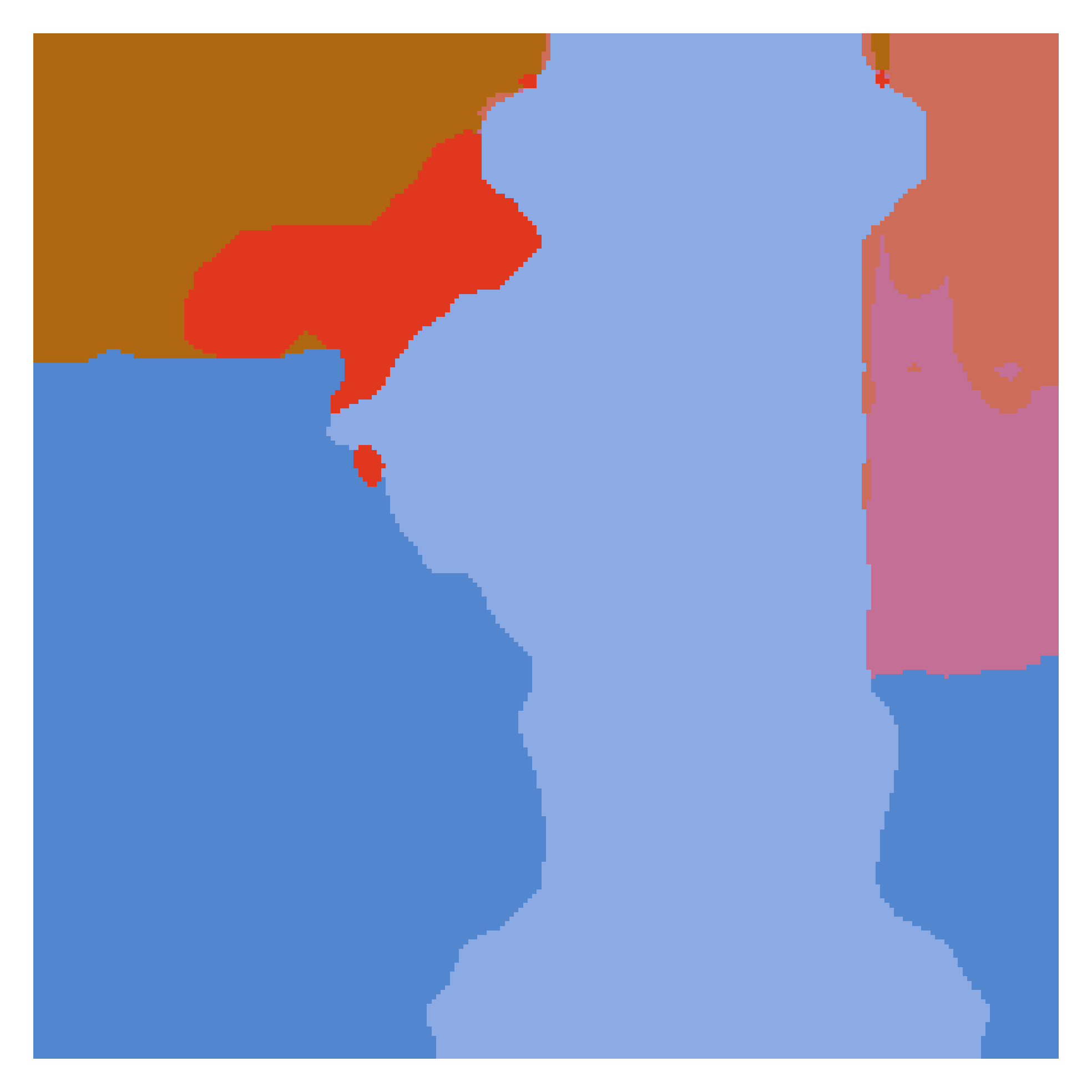} \\[-2pt]
    \includegraphics[width=0.11\textwidth]{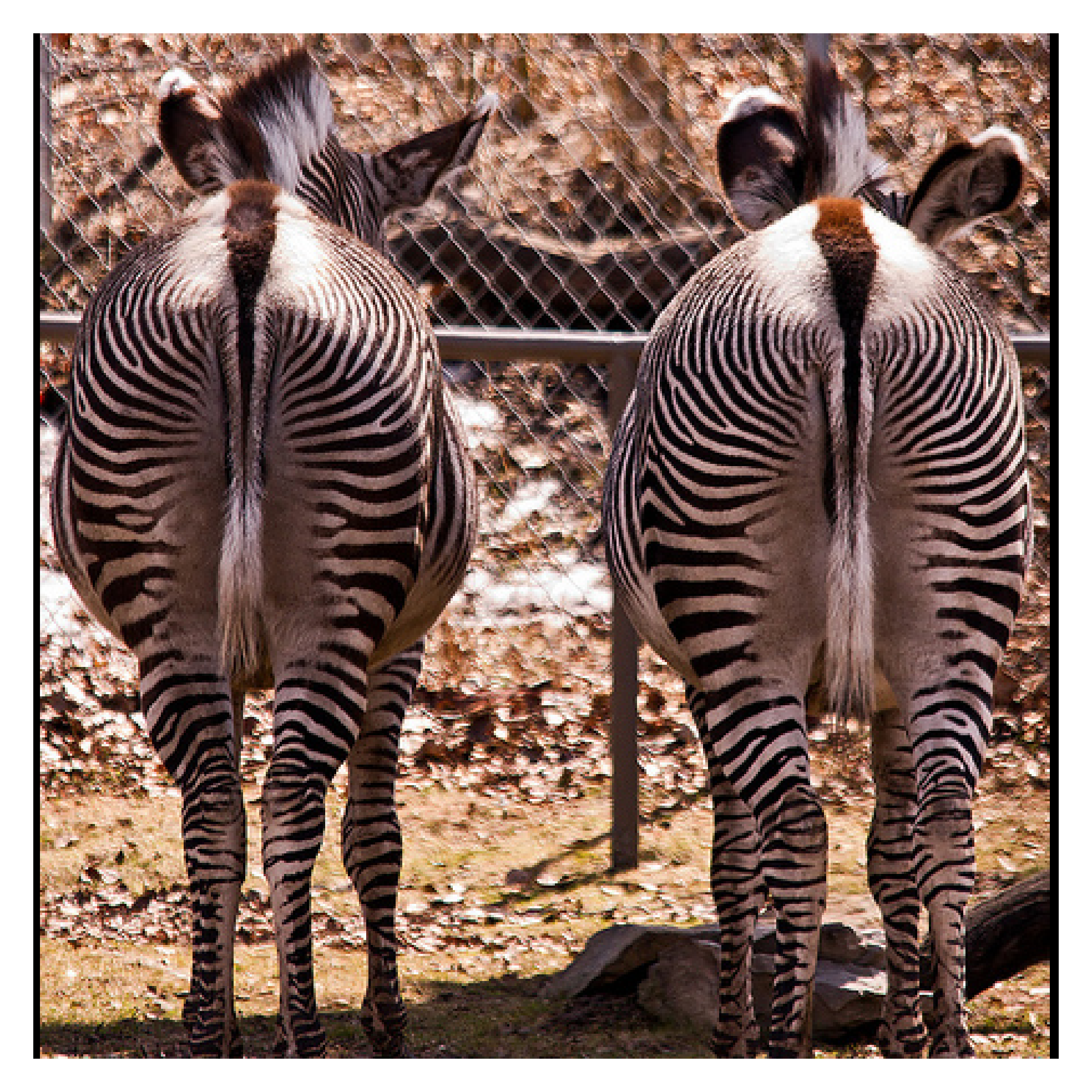} &
    \includegraphics[width=0.11\textwidth]{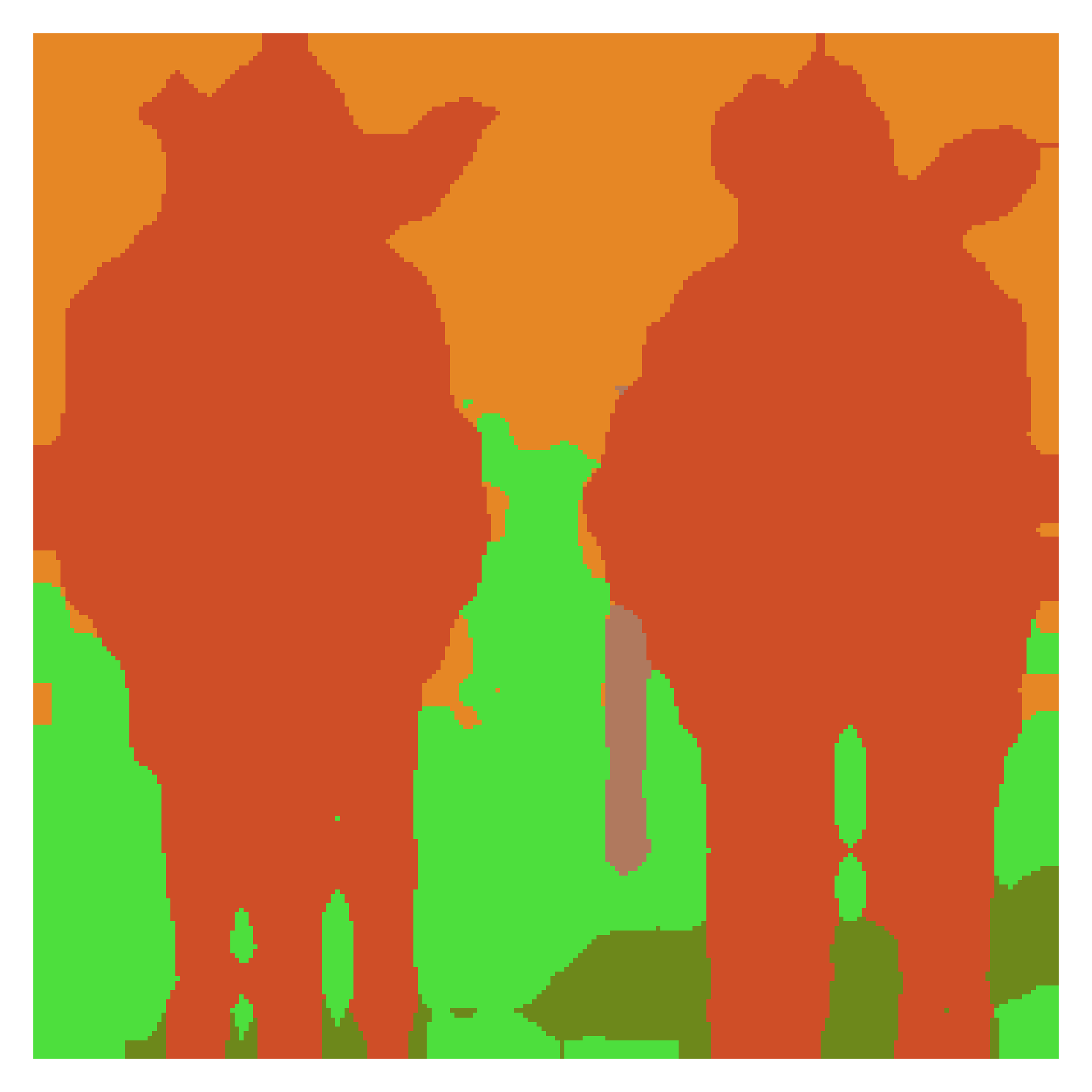} &
    \includegraphics[width=0.11\textwidth]{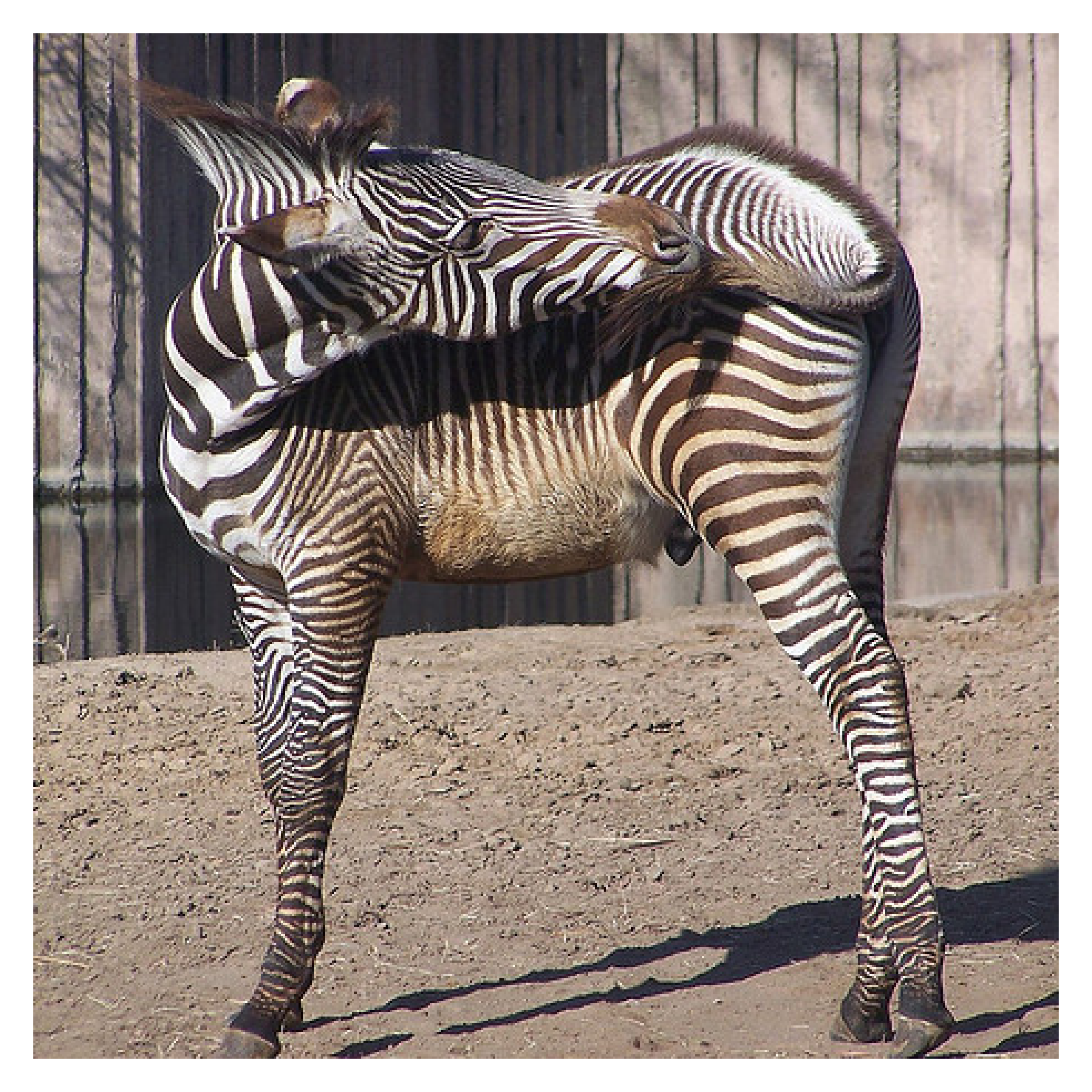} &
    \includegraphics[width=0.11\textwidth]{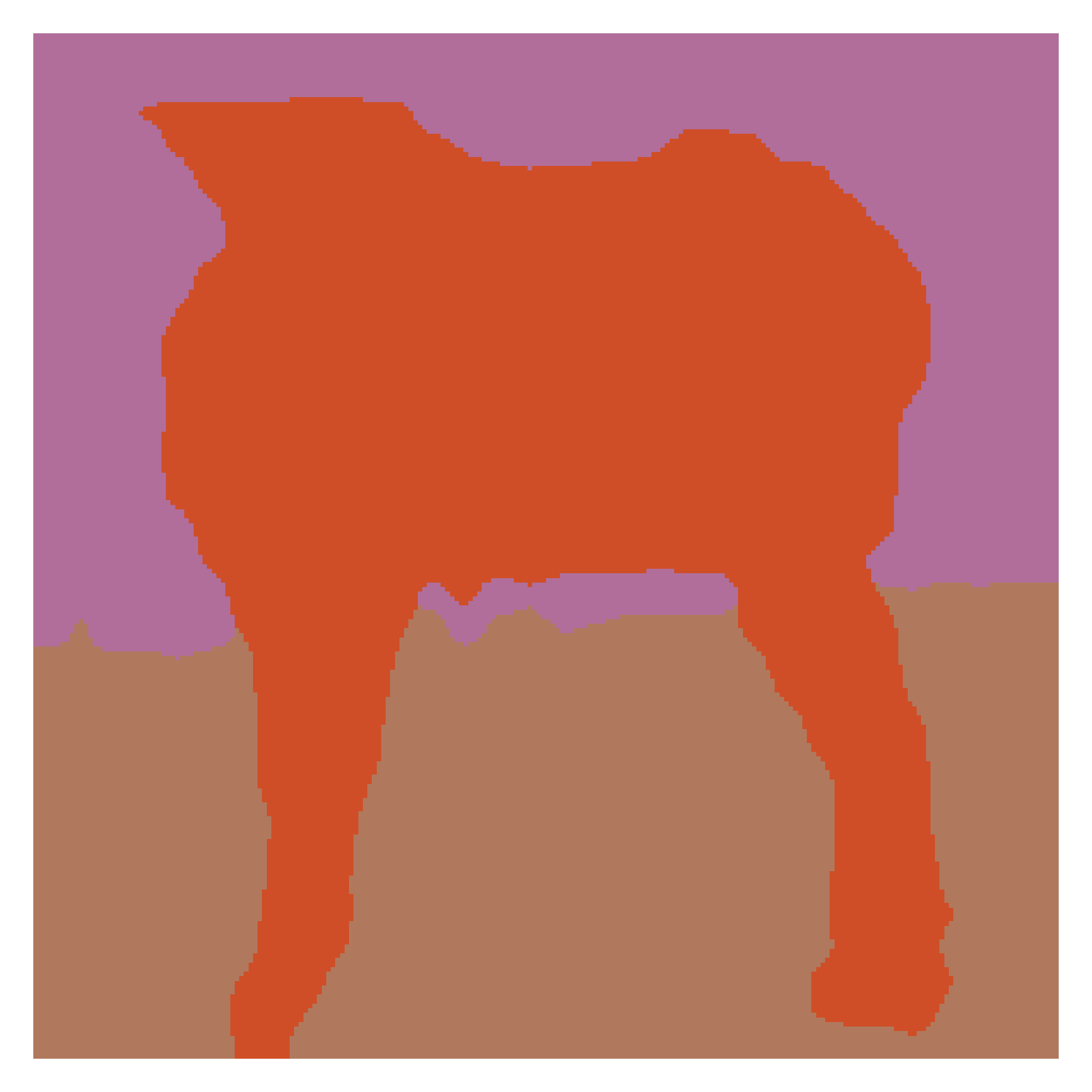} \\[-1pt]
    {\scriptsize (a) Query} & {\scriptsize (b)Pseudo labels} & {\scriptsize (c) Positive} & {\scriptsize (d) Pseudo labels} \\
  \end{tabular}
  \vspace{-8pt}

 \caption{\textbf{Examples of automatically constructed in-context scene understanding tasks.} Each row shows a query image and its corresponding positive support example. (a) and (b) display the query image and its pseudo segmentation labels, while (c) and (d) show the positive support image and its pseudo segmentation labels. Despite being generated in a fully unsupervised manner, the segmentation masks for salient objects are highly accurate, closely matching the actual objects. 
  In addition, the query and positive image pairs share a common object with the same pseudo-label. During post-training with our \ours method, the model predicts the query image's segmentation using the positive example as a reference, along with distractor support examples (randomly sampled from other images in the mini-batch, not shown here for brevity).}
  \label{fig:pseudolabel_labels}
  \vspace{-10pt}
\end{figure}

\subsection{Automatic Dense In-Context Task Construction} \label{sec:task_construction}

To apply our dense in-context post-training approach to unlabeled data, we automatically generate in-context pseudo-tasks with spatially coherent segmentation labels. This involves two steps: (1) generating pseudo-segmentation maps for dataset images and (2) selecting query and support examples to form each in-context task. These steps create meaningful and challenging tasks, enabling the model to learn transferable in-context visual learning skills. Examples of constructed pseudo-tasks are shown in \cref{fig:pseudolabel_labels}.

\parag{Generating pseudo semantic segmentation labels.} 
We generate pseudo-segmentation maps by first dividing each image into non-overlapping class-agnostic segments and then assigning a pseudo-label to each segment.

For the first step, we use DiffCut \cite{diffcut}, a training-free zero-shot image segmentation method. DiffCut leverages features from a pretrained diffusion model, SSD-1B \cite{ssd1b}, along with a recursive graph partitioning algorithm to produce fine-grained segmentation maps. These maps, called DiffCut masks, are generated and stored for the entire pretraining dataset. Since DiffCut is unsupervised, our pretraining approach remains fully annotation-free.

For the second step, we assign pseudo-labels to each segment using the self-supervised DINOv2R feature encoder. We compute the mean DINOv2R feature for each DiffCut mask by pooling dense features within the mask region. To group visually similar segments, we apply K-means clustering on these pooled features\footnote{For COCO~\cite{coco}, clustering is performed on the full dataset; for ImageNet~\cite{imagenet}, clustering is performed on a random subset of 200,000 training images for efficiency.}. The resulting clusters serve as pseudo-classes for annotating the DiffCut masks. To assign a pseudo-class to each DiffCut mask, we first assign each pixel-wise DINOv2R feature the cluster ID of its closest K-means centroid. Then, for each DiffCut mask, we perform majority voting on all pixels within the mask and assign the most frequent cluster ID. This ensures robust and consistent pseudo-labels, even with noisy features.

\parag{Selecting examples for in-context tasks.} 
To generate in-context tasks, we pair each query image with one ``positive" support image and $K$-1 ``distractor" support images.

To find positive images sharing similar visual content, we use DINOv2R global image representations. For each image, we retrieve five nearest neighbors using DINOv2R global features. This yields five positive pairs per image. We then retain pairs where the query and positive-support images share a common pseudo-class occupying more than $5\%$ of each image's area. This ensures semantically meaningful positive pairs. The final refined list contains all possible query-positive support pairs for training.

For the $K$-1 ``distractor" examples, we randomly sample images from the current mini-batch during training. Including distractors encourages the model to distinguish relevant from irrelevant content, improving feature discriminability.

\section{Experiments}

\subsection{Experimental Setup} \label{sec:setup}
\begin{table*}[ht!]
    \small
    \centering
    \begin{tabular}{llcccccccc} 
        \toprule
        \multirow{2}{*}{\textbf{Method}}& \multirow{2}{*}{\textbf{Backbone}} & \multicolumn{4}{c}{\textbf{PascalVOC}} & \multicolumn{4}{c}{\textbf{ADE20K}} \\
        & & $\frac{1}{1}$ & $\frac{1}{8}$ & $\frac{1}{64}$ & $\frac{1}{128}$ & $\frac{1}{1}$ & $\frac{1}{8}$ & $\frac{1}{64}$ & $\frac{1}{128}$\\
        \midrule 
        DINO$^\dagger$~\cite{caron2021emerging}      & ViT-S/16 & 48.7 & 41.3 & 30.5 & 26.4 & 17.9 & 15.0 & 11.0 & 9.5\\
        SelfPatch$^\dagger$~\cite{selfpatch} & ViT-S/16 & 50.8 & 43.2 & 32.6 & 28.4 & 17.7 & 14.7 & 10.9 & 10.0\\
        CroC$^\dagger$~\cite{stegmuller2023croc}     & ViT-S/16 & 60.5 & 53.8 & 41.8 & 34.0 & 17.3 & 15.2 & 10.8 & \,\,\,8.7\\
        TimeT$^\dagger$~\cite{salehi2023time}     & ViT-S/16 & 62.3 & 55.2 & 43.8 & 38.1 & 23.2 & 18.9 & 14.1 & 12.1\\
        Leopart$^\dagger$~\cite{ziegler2022self}   & ViT-S/16 & 64.5 & 58.4 & 49.7 & 44.6 & 23.9 & 19.6 & 14.8 & 12.9\\
        CriBO$^\dagger$~\cite{cribo}     & ViT-S/16 & 72.4 & 66.9 & 59.9 & 53.9 & 26.6 & 22.7 & 17.3 & 14.6\\
        DINOv2R$^\ddagger$~\cite{darcet2024vision}  & ViT-S/14 & 79.4 & 75.2& 67.7 & 60.7 & 39.3 & 33.2 & 24.9 & 22.6\\
        NeCo$^\ddagger$~\cite{neco}     & ViT-S/14 & \textbf{81.0} & 77.3 & \textbf{71.5} & 65.8 & 38.9 & 32.5 & 24.1 & 21.9\\
        \rowcolor{baselinecolor} \ours (ours)$^\ddagger$ & ViT-S/14 & \textbf{81.0}  & \textbf{77.7} & 71.4  & \textbf{ 65.9} & \textbf{39.7} & \textbf{33.7} & \textbf{25.6} & \textbf{23.2}\\
        \midrule
        DINO$^\dagger$~\cite{caron2021emerging}        & ViT-B/16& 57.3 & 49.8 & 37.7 & 33.1 & 21.5 & 18.2 & 13.5 & 11.5\\
        Leopart$^\dagger$~\cite{ziegler2022self}     & ViT-B/16& 69.5 & 63.1 & 54.7 & 50.1 & 26.7 & 21.8 & 16.8 & 14.6\\
        Hummingbird$^\dagger$~\cite{hummingbird} & ViT-B/16& 71.8 & 64.3 & 57.2 & 50.5 & 29.6 & 22.3 & 15.1 & 11.7\\
        CriBO$^\dagger$~\cite{cribo}       & ViT-B/16& 74.2 & 69.2 & 61.8 & 55.9 & 28.4 & 24.4 & 18.4 & 15.9\\
        DINOv2R$^\ddagger$~\cite{darcet2024vision}    & ViT-B/14 & 79.0 & 75.3 & 67.6 & 60.3 & 40.8 & 35.3 & 27.3 & 24.9\\
        NeCo$^\ddagger$~\cite{neco}       & ViT-B/14 & \textbf{82.4} & 78.7 & 71.7 & 65.0 & 41.2 & 35.2 & 27.2 & 25.1\\
        \rowcolor{baselinecolor} \ours (ours)$^\ddagger$ & ViT-B/14 & 82.1 & \textbf{79.6} & \textbf{75.1} & \textbf{70.1} & \textbf{42.6} & \textbf{36.8} & \textbf{29.4} & \textbf{27.0}\\
        \bottomrule
    \end{tabular}
    \vspace{-8pt}
    \caption{\textbf{In-context scene understanding with few training examples.}  
    Dense nearest neighbor retrieval performance on ADE20K and PascalVOC datasets, evaluated using the mIoU metric across varying proportions of training data. The fractions $\frac{x}{y}$ below the dataset names indicate the proportion of training data used. Results marked with $\dagger$ are from NeCo \cite{neco}, while those marked with $\ddagger$ are our own.}
    \label{tab:ic_segm_lowdata}
    \vspace{-8pt}
\end{table*}
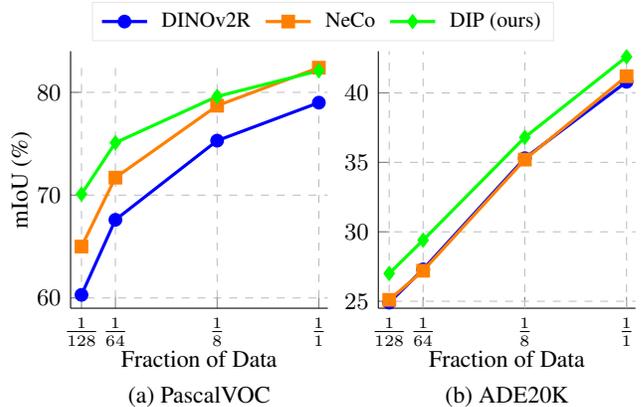
\begin{figure}[t] 
    \centering
    \setlength{\tabcolsep}{0pt} 
    \begin{tabular}{@{}c@{}c@{}} 
    \multicolumn{2}{c}
    {
    \begin{tikzpicture}
        \centering
        \begin{customlegend}[legend columns=3, legend style={align=center, column sep=.7ex, nodes={scale=0.9, transform shape}, draw=white!90!black}, legend entries={ \small DINOv2R, \small NeCo, \small DIP (ours)}]
            \addlegendimage{color=blue, mark=*, line width=1.2pt}
            \addlegendimage{color=orange, mark=square*, line width=1.2pt}
            \addlegendimage{color=green, mark=diamond*, line width=1.2pt}
        \end{customlegend}
    \end{tikzpicture}
    }
    \\
    \hspace{-0.05\linewidth}
        \begin{tikzpicture}
            \tikzstyle{every node}=[font=\small]
            \begin{axis}[
                width=0.6\linewidth,
                height=0.6\linewidth, 
                xlabel={Fraction of Data},
                ylabel={mIoU (\%)},
                ylabel style={shift={(0pt,-5pt)}}, 
                xlabel style={shift={(-0pt,5pt)}}, 
                font=\small,  
                xmode=log,
                log basis x=2, 
                xtick={1/128, 1/64, 1/8, 1},
                xticklabels={$\frac{1}{128}$, $\frac{1}{64}$, $\frac{1}{8}$, $\frac{1}{1}$},
                ymajorgrids=true, 
                grid style={dashed, gray!50},
                xmin=1/160, xmax=1.2, 
                ymin=59, ymax=84, 
                legend cell align={left},
                axis y line*=left,
                axis x line*=bottom,
                tick label style={font=\small}, 
                label style={font=\small}, 
                title={(a) PascalVOC},
                title style={font=\small},
                title style={yshift=-32.ex,},    
            ]
                \addplot[
                    color=blue,
                    mark=*,
                    line width=1.2pt,
                ] coordinates {
                    (1/128, 60.3)
                    (1/64, 67.6)
                    (1/8, 75.3)
                    (1, 79.0)
                };

                \addplot[
                    color=orange,
                    mark=square*,
                    line width=1.2pt,
                ] coordinates {
                    (1/128, 65.0)
                    (1/64, 71.7)
                    (1/8, 78.7)
                    (1, 82.4)
                };

                \addplot[
                    color=green,
                    mark=diamond*,
                    line width=1.2pt,
                ] coordinates {
                    (1/128, 70.1)
                    (1/64, 75.1)
                    (1/8, 79.6)
                    (1, 82.1)
                };

                \foreach \x in {1/128, 1/64, 1/8, 1} {
                    \edef\temp{\noexpand\draw[dashed, gray!50] (axis cs:\x,60) -- (axis cs:\x,83);}
                    \temp
                }
            \end{axis}
        \end{tikzpicture}
    &
        \begin{tikzpicture}
        \tikzstyle{every node}=[font=\small]
            \begin{axis}[
                width=0.6\linewidth,
                height=0.6\linewidth, 
                xlabel={Fraction of Data},
                xlabel style={shift={(-0pt,5pt)}}, 
                font=\small,  
                xmode=log,
                log basis x=2, 
                xtick={1/128, 1/64, 1/8, 1},
                xticklabels={$\frac{1}{128}$, $\frac{1}{64}$, $\frac{1}{8}$, $\frac{1}{1}$},
                ymajorgrids=true, 
                grid style={dashed, gray!50},
                xmin=1/160, xmax=1.2, 
                ymin=24.5, ymax=43, 
                legend cell align={left},
                axis y line*=left,
                axis x line*=bottom,
                tick label style={font=\small},
                label style={font=\small},
                title={(b) ADE20K},
                title style={font=\small},
                title style={yshift=-32.ex,},                
            ]
                \addplot[
                    color=blue,
                    mark=*,
                    line width=1.2pt,
                ] coordinates {
                    (1/128, 24.9)
                    (1/64, 27.3)
                    (1/8, 35.3)
                    (1, 40.8)
                };

                \addplot[
                    color=orange,
                    mark=square*,
                    line width=1.2pt,
                ] coordinates {
                    (1/128, 25.1)
                    (1/64, 27.2)
                    (1/8, 35.2)
                    (1, 41.2)
                };

                \addplot[
                    color=green,
                    mark=diamond*,
                    line width=1.2pt,
                ] coordinates {
                    (1/128, 27.0)
                    (1/64, 29.4)
                    (1/8, 36.8)
                    (1, 42.6)
                };

                \foreach \x in {1/128, 1/64, 1/8, 1} {
                    \edef\temp{\noexpand\draw[dashed, gray!50] (axis cs:\x,24.5) -- (axis cs:\x,43);}
                    \temp
                }
            \end{axis}
        \end{tikzpicture}
        \hspace*{-0.05\linewidth} 
    \end{tabular}
    \vspace{-10pt}
    \caption{\textbf{In-context scene understanding in low-shot regimes.}
mIoU results with ViT-B/14 versus training data size.}
    \label{fig:ic_segm_function_of_data}
    \vspace{-10pt}
\end{figure}

\parag{Compared methods.}  
We compare our method with state-of-the-art unsupervised learning approaches, including DINO~\cite{caron2021emerging}, Leopart~\cite{ziegler2022self}, TimeT~\cite{salehi2023time}, iBOT~\cite{zhou2022ibot}, CrOC~\cite{stegmuller2023croc}, CrIBO~\cite{cribo}, DINOv2R~\cite{dinov2,darcet2024vision}, and NeCo~\cite{neco}. NeCo, a recent method (ICLR’25), is the most related to ours, as it also post-trains DINOv2R for dense representation learning in in-context scene understanding tasks. However, our post-training methodology differs significantly: we explicitly train on in-context pseudo-tasks relevant to our target applications, while NeCo adopts a self-distillation-based approach.

\parag{Training setup.}  
We apply our unsupervised post-training to DINOv2~\cite{dinov2} with registers~\cite{darcet2024vision} (DINOv2R), using ViT-S/14 and ViT-B/14 models. In-context pseudo-tasks are generated using DiffCut and the base model (e.g., DINOv2R ViT-B/14 is used to generate pseudo-tasks for post-training DINOv2R ViT-B/14; see \cref{sec:task_construction}). Except otherwise stated, post-training is on COCO for 5 epochs with 1,000 pseudo-classes. We also explore post-training CLIP and MAE models with ViT-B/16 (using pseudo-labels from DiffCut and DINOv2R ViT-B/14), as well as post-training on ImageNet. 
Our post-training is efficient, requiring 17h52min on V100 (ViT-B) and 8h37min on A100 (ViT-S). Pseudo-label generation for COCO, which is done only once and offline, requires 17h on a
V100 GPU. More implementation details in \cref{sec:imp-details}.

\parag{Evaluation setup.}  
To assess our representations' generality, we evaluate them on diverse tasks and datasets. These include semantic segmentation on PascalVOC~\cite{Pascalvoc}, Pascal-Context~\cite{Pascalcontext}, ADE20K~\cite{ade20k}, Cityscapes~\cite{cityscapes}, COCO~\cite{coco} (mIoU$\uparrow$), and monocular depth prediction on NYUv2~\cite{nyu} (using RMSE$\downarrow$). COCO is in-domain, as its training data is used for post-training, while the others are out-of-domain. These datasets test generalization to domain shifts between post-training and downstream tasks.  
We also evaluate robustness to intra-task domain shifts using the CS$\rightarrow$ACDC~\cite{acdc} setting, where Cityscapes (CS) provides the support set and ACDC provides the query images for retrieval-based segmentation. While both datasets contain autonomous driving images, ACDC introduces challenging conditions like snow, night, fog, and rain, unlike Cityscapes' clear-weather daylight images.

All results reported in this paper are produced using our implementation unless marked with $\dagger$. For DINOv2R and NeCo, we evaluate the publicly available pretrained model checkpoints to ensure a fair comparison.

\begin{table*}[t]
    \small
    \centering
    \centering
    \begin{tabular}{l l l l l l l l c} 
        \toprule
        Method & Backbone & ADE20K & PascalVOC &  Pascal-Context & Cityscapes & CS$\rightarrow$ACDC & COCO & Avg. Delta\\
        \midrule
        DINOv2R~\cite{darcet2024vision}  & ViT-S/14 & 39.3 & 79.4 & 48.0 & 55.6 & \textbf{47.4} & 72.6 & \\
        NeCo~\cite{neco}     & ViT-S/14 & 38.9 (\textcolor{red}{$-$0.4}) & \textbf{81.0} (\textbf{\textcolor{blue}{$+$1.6}}) & 49.4 (\textcolor{blue}{$+$1.4}) & 53.9 (\textcolor{red}{$-$1.7}) & 47.2 (\textcolor{red}{$-$0.2}) & \textbf{74.3} (\textbf{\textcolor{blue}{$+$1.7}}) & \textcolor{blue}{$+$0.40} \\
        \rowcolor{baselinecolor} \ours (ours) & ViT-S/14 & \textbf{39.7} (\textbf{\textcolor{blue}{$+$0.4}}) & \textbf{81.0} (\textbf{\textcolor{blue}{$+$1.6}})  & \textbf{49.5} (\textbf{\textcolor{blue}{$+$1.5}}) & \textbf{55.8} (\textbf{\textcolor{blue}{$+$0.2}}) & \textbf{47.4} ($+$0.0) & 74.0 (\textcolor{blue}{$+$1.4}) & \textbf{\textcolor{blue}{$+$0.85}}\\
        \midrule
        DINOv2R~\cite{darcet2024vision}  & ViT-B/14 & 40.8 & 79.0 & 49.0 & 58.4 & 50.5 & 72.9 & \\
        NeCo~\cite{neco}     & ViT-B/14 & 41.2 (\textcolor{blue}{$+$0.4}) & \textbf{82.4} (\textbf{\textcolor{blue}{$+$2.4}}) & 51.2 (\textcolor{blue}{$+$2.2}) & 57.9 (\textcolor{red}{$-$0.5}) & 51.5 (\textcolor{blue}{$+$1.0}) & 75.4 (\textcolor{blue}{$+$2.5}) & \textcolor{blue}{$+$1.50} \\
        \rowcolor{baselinecolor} \ours (ours) & ViT-B/14 & \textbf{42.6} (\textbf{\textcolor{blue}{$+$1.8}}) & 82.1 (\textcolor{blue}{$+$2.1}) & \textbf{51.5} (\textcolor{blue}{\textbf{$+$2.5}}) & \textbf{59.5} (\textbf{\textcolor{blue}{$+$1.1}}) & \textbf{52.1} (\textbf{\textcolor{blue}{$+$1.6}}) & \textbf{76.0} (\textbf{\textcolor{blue}{$+$3.1}}) & \textbf{\textcolor{blue}{$+$2.00}}\\
        \bottomrule
    \end{tabular}
    \vspace{-8pt}
    \caption{\textbf{In-context scene understanding benchmark.}
    Dense nearest neighbor retrieval performance for semantic segmentation on six scene-centric datasets: ADE20K, PascalVOC, Pascal-Context, Cityscapes, CS$\rightarrow$ACDC, and COCO. The first five are out-of-domain, while COCO is in-domain (used for post-training). Performance is measured using the mIoU metric. For NeCo and our \ours, which post-train DINOv2R, improvements over DINOv2R are shown in parentheses. The ``Avg. Delta'' column reports the average improvement.}
    \label{tab:ic_segm_many}
    \vspace{-10pt}
\end{table*}

\subsection{Retrieval-based Scene Understanding}

\parag{Comparison with state-of-the-art in low-shot regimes.} 
In \cref{tab:ic_segm_lowdata}, we compare our \ours approach with prior state-of-the-art methods for dense image representation learning on retrieval-based semantic segmentation using the Pascal VOC and ADE20K datasets. We evaluate both full-data ($\frac{1}{1}$) and low-shot regimes ($\frac{1}{8}$, $\frac{1}{64}$, and $\frac{1}{128}$), where only a fraction of the training examples is used in the support set.

Our \ours approach consistently outperforms DINOv2R, with the performance gap widening as the number of training examples decreases, particularly for the PascalVOC dataset (see \cref{fig:ic_segm_function_of_data}). Notably, \ours achieves superior results compared to prior work in most settings for both ViT-S and ViT-B, while remaining competitive in others.

\begin{table}[t!]
    \small
    \centering
    \begin{tabular}{l c c c}
        \toprule
        Method & DINOv2R & NeCo & \baseline{\ours (ours)}\\
        \midrule
        RMSE$\downarrow$ & .771 & .769 & \baseline{\textbf{.756}}\\
        \bottomrule
    \end{tabular}
    \vspace{-8pt}
    \caption{\textbf{In-context monocular depth prediction on NYUv2 dataset \cite{silberman2012indoor}.}
    RMSE scores (lower is better) scaled by 10 for readability reasons. All methods use ViT-S/14.
    } 
    \label{tab:nyudepth}
    \vspace{-5pt}
\end{table}

\parag{Comparison with DINOv2R and NeCo.}  
We compare our \ours method with DINOv2R, the base model for our post-training, and NeCo, a recent method that also post-trains DINOv2R. In \cref{tab:ic_segm_many} we evaluate these methods on in-context semantic segmentation tasks across six datasets.

Our approach shows consistent improvements over DINOv2R on all datasets, both in-domain (COCO) and out-of-domain (ADE20K, PascalVOC, Pascal-Context, and Cityscapes), highlighting the generalization capability of our representations. Unlike NeCo, our \ours consistently improves over DINOv2R, achieving higher mIoU and average performance improvements (see ``Avg. Delta" in \cref{tab:ic_segm_many}).

\parag{Intra-task domain shift.}  
In \cref{tab:ic_segm_many}, the CS$\rightarrow$ACDC setting evaluates the robustness of the learned representations to domain shifts between the support set and query images during the downstream stage, as explained in \cref{sec:setup}. Although our post-training \ours method is not specifically designed to improve this type of robustness, it still achieves a 1.6 mIoU improvement over DINOv2R for ViT-B.

\parag{Monocular depth prediction.}
In \cref{tab:nyudepth}, we report in-context depth results on NYUv2 (RMSE). Both \ours and NeCo improve over DINOv2R, with \ours achieving the best RMSE. This is notable, as our in-context pseudo-tasks are designed for semantic segmentation, yet the representations generalize well to depth prediction.

\begin{table}[t!]
    \small
    \centering
    \begin{tabular}{l l l l}
    \toprule
    Method & Backbone & PascalVOC & ADE20K\\
    \midrule
    DINOv2R~\cite{darcet2024vision} & ViT-B/14 & 79.0 & 40.8\\
    +\baseline{\ours}  & ViT-B/14 & \baseline{82.1  (\textbf{\textcolor{blue}{$+$2.1}})} & \baseline{42.6 (\textbf{\textcolor{blue}{$+$1.8})}}\\
    \midrule
    CLIP~\cite{clip} & ViT-B/16 & 71.8 & 29.0 \\
    +\baseline{ \ours }&  ViT-B/16 & \baseline{73.8  (\textbf{\textcolor{blue}{$+$2.0}})} & \baseline{30.1  (\textbf{\textcolor{blue}{$+$1.1}})} \\
    \midrule
    MAE~\cite{he2022masked} & ViT-B/16 & 13.9 & 5.1 \\
    +\baseline{ \ours }&  ViT-B/16 & \baseline{47.3 (\textbf{\textcolor{blue}{$+$33.4}})} & \baseline{11.8 (\textbf{\textcolor{blue}{$+$6.7}})} \\
    \bottomrule
    \end{tabular}
    \vspace{-8pt}
    \caption{\textbf{Post-training with other base models.}
    Dense nearest neighbor retrieval performance (mIoU) for semantic segmentation on ADE20K and PascalVOC, using DINOv2R, CLIP, and MAE as base models, before and after post-training.}
    \label{tab:base_models}
    \vspace{-10pt}
\end{table}

\parag{Post-training other base models.}  
In \cref{tab:base_models}, we evaluate in-context segmentation on ADE20K and PascalVOC using CLIP~\cite{clip} and MAE~\cite{he2022masked} as base models, before and after \ours post-training. Our method consistently improves performance across all tested models, showing its versatility. MAE improves by \textbf{$+33.4$} mIoU on PascalVOC. These gains transform MAE from weak to improved at in-context segmentation, highlighting our approach’s effectiveness in enhancing dense features.

\begin{table}[t!]
    \small
    \centering
 
    \begin{tabular}{l l c c c}
    \toprule
    \textbf{Method} & \textbf{Backbone} & \textbf{PascalVOC} & \textbf{ADE20K} \\ \midrule
    
    SD \cite{ssd1b} & SSD-1B & 59.4& 24.4 \\
    \midrule
    SAM \cite{kirillov2023segment}& ViT-B/16& 32.8 & 12.9 \\
    DINOv2R \cite{darcet2024vision}& ViT-B/14& 79.0& 40.8 \\
    RADIOv2.5 \cite{radio} &ViT-B/16& 81.3& 42.1 \\
    \baseline{\ours (ours)} & \baseline{ViT-B/14}& \baseline{\textbf{82.1}}  & \baseline{\textbf{42.6}} \\
    \bottomrule
    \end{tabular}
    \caption{\textbf{Comparison with Superised Baselines}. Dense nearest neighbor retrieval performance (mIoU) for semantic segmentation on PascalVOC and ADE20K.} 
    \label{tab:sup_baselines}
     \vspace{-10pt}
\end{table}

\parag{Comparison with supervised baselines}
To automatically generate pseudo-segmentations, our post-training approach leverages Stable Diffusion (SD) features \cite{ssd1b}, trained on weakly annotated (internet-scraped) image-caption pairs. In \cref{tab:sup_baselines}, we compare \ours to SD and supervised encoder features (SAM \cite{kirillov2023segment} and RADIOv2.5 \cite{radio}). Crucially, SAM requires \emph{stronger supervision} (manual segmentation masks) than SD, while RADIOv2.5 distills multiple model types (SAM, DINOv2, and CLIP-like features). Thus, both baselines leverage more supervision than our method. Additionally, RADIOv2.5 requires significantly more training compute. Despite these advantages, \cref{tab:sup_baselines} shows \ours surpasses RADIOv2.5 in in-context semantic segmentation, confirming our method's effectiveness. SAM underperforms in this setting, consistent with prior findings \cite{kerssies2024benchmarking}. Importantly, \ours outperforms the SD features it uses for pseudo-segmentation.

Additionally, we report linear segmentation results with consistent gains over DINOv2R and competitive results compared to NeCo in \cref{tab:lin_segm} in supplementary.

\subsection{Ablations}  
\begin{table*}[ht!]
\vspace{-.2em}
\centering
\subfloat[
\textbf{In-context pretraining} is more effective. 
\label{tab:ablation_task}
]{
\begin{minipage}{0.35\linewidth}{\begin{center}
\tablestyle{4pt}{1.05}
\begin{tabular}{l c c c}
\toprule
Objective & PascalVOC & ADE20K & NYUv2$\downarrow$ \\\midrule
\baseline{In-context} & \baseline{\textbf{81.0}}  & \baseline{39.7} & \baseline{\textbf{.756}}\\
Direct.      & 79.9  &  \textbf{39.9}  & .776  \\
\bottomrule
\end{tabular}
\end{center}}\end{minipage}
}
\subfloat[
\textbf{Distractor examples} are important.
\label{tab:ablation_distractor}
]
{
\begin{minipage}{0.3\linewidth}{\begin{center}
\tablestyle{4pt}{1.05}
\begin{tabular}{c c c}
\toprule
Distractor & PascalVOC & ADE20K \\\midrule
\baseline{\cmark} & \baseline{\textbf{81.0}}  & \baseline{\textbf{39.7}} \\
\xmark            & 78.5  &  38.4   \\
\bottomrule
\end{tabular}
\end{center}}\end{minipage}
}
\subfloat[
\textbf{Constructing positive examples.}
\label{tab:ablation_positives}
]{
\begin{minipage}{0.34\linewidth}{\begin{center}
\tablestyle{4pt}{1.05}
\begin{tabular}{l c c}
\toprule
Positive example & PascalVOC & ADE20K \\\midrule
\baseline{Nearest neighbor} & \baseline{\textbf{81.0}}  & \baseline{\textbf{39.7}} \\
Two random crops            & 79.5  &  39.4   \\
\bottomrule
\end{tabular}
\end{center}}\end{minipage}
}
\\
\centering
\vspace{.3em}
\subfloat[
\textbf{Pseudo-labels generation.}
\label{tab:ablation_pseudo_labels}
]{
\begin{minipage}{0.32\linewidth}{\begin{center}
\tablestyle{4pt}{1.05}
\begin{tabular}{l c c}
\toprule
Segm. Labels & PascalVOC & ADE20K \\\midrule
\baseline{Pseudo-labels w/ DiffCut} & \baseline{81.0}  & \baseline{\textbf{39.7}} \\
Pseudo-labels w/o DiffCut        & 59.1  &  25.5   \\
\midrule
Ground truth labels        & \textbf{81.9}  & \textbf{39.8}     \\
\bottomrule
\end{tabular}
\end{center}}\end{minipage}
}
\hspace{2em}
\subfloat[
\textbf{Number of pseudo-classes.}
\label{tab:ablation_num_pseudo_classes}
]{
\begin{minipage}{0.28\linewidth}{\begin{center}
\tablestyle{4pt}{1.05}
\begin{tabular}{r c c}
\toprule
\#Pseudo-classes & PascalVOC & ADE20K \\\midrule
300 &80.8 & 39.5\\
500 & 80.9& 39.7\\
\baseline{1000} & \baseline{\textbf{81.0}}  & \baseline{\textbf{39.7}} \\
2000 & 80.8&39.8 \\
\bottomrule
\end{tabular}
\end{center}}\end{minipage}
}
\hspace{2em}
\subfloat[
\textbf{Post-training dataset.}
\label{tab:ablation_dataset}
]{
\begin{minipage}{0.28\linewidth}{\begin{center}
\tablestyle{4pt}{1.05}
\begin{tabular}{l c c}
\toprule
Dataset & PascalVOC & ADE20K \\\midrule
\baseline{COCO} & \baseline{\textbf{81.0}}  & \baseline{\textbf{39.7}} \\
ImageNet        & 80.8  &  39.6   \\
\bottomrule
\end{tabular}
\end{center}}\end{minipage}
}

\vspace{-8pt}
\caption{\textbf{Ablation Study of \ours.} 
Default settings in \colorbox{baselinecolor}{light blue}. 
``Two random crops'' in (c) refers to the case where the query and positive image are constructed as two random crops of the same original image.}
\label{tab:ablations} 
\vspace{-10pt}
\end{table*}

In \cref{tab:ablations}, we evaluate key design choices of \ours approach. 

\parag{In-context vs Direct dense prediction (\cref{tab:ablation_task}).}  
We compare our in-context dense prediction objective with direct dense prediction, where the model predicts pseudo semantic segmentation maps using a classification head. Our in-context approach significantly outperforms the direct method on PascalVOC and reduces RMSE on NYUv2 depth prediction, while both perform similarly on ADE20K. 

\parag{Impact of ``distractor" examples (\cref{tab:ablation_distractor}).}
Removing the ``distractor" support examples significantly reduces performance, as the absence of ``distractors" simplifies the post-training task, limiting the model's ability to learn discriminative dense representations.

\parag{Construction of positive examples (\cref{tab:ablation_positives}).} 
We compare two strategies for creating positive examples: (1) retrieving nearest neighbors using DINOv2R image-wise features and (2) using two random crops from the same image. The nearest neighbor strategy outperforms random crops, validating our design choice and showing that our method avoids reliance on complex augmentations common in self-distillation and contrastive approaches. 

\parag{Impact of DiffCut (\cref{tab:ablation_pseudo_labels}).} 
We assess the role of DiffCut \cite{diffcut}, which generates class-agnostic segments for pseudo segmentation maps before K-means clustering. Compared to direct K-means on DINOv2R features, DiffCut significantly improves performance, demonstrating its effectiveness as a training-free, unsupervised method leveraging Stable Diffusion. Furthermore, our pseudo labels achieve results nearly matching ground-truth semantic segmentation, underscoring their quality.

\parag{Impact of number of pseudo classes (\cref{tab:ablation_num_pseudo_classes}).} 
We study the effect of pseudo-class count (K-means clusters) for pseudo-label generation. Our approach is robust to this hyperparameter, with stable performance across values.

\parag{Post-training data: scene-centric vs object-centric (\cref{tab:ablation_dataset}).} 
We compare \ours post-trained on scene-centric COCO vs. object-centric ImageNet. Results are comparable, COCO slightly outperforms ImageNet. This indicates that our method does not depend on human-curated, object-centric data and performs better on scene-centric data, easier to collect at scale.

\subsection{Qualitative results}

\begin{figure}[t!]
    \small
    \centering
    \begin{tabular}{c@{\hspace{0.1cm}}c@{\hspace{0.1cm}}c@{\hspace{0.1cm}}c}
       \includegraphics[width=0.235\linewidth,height=3cm,keepaspectratio]{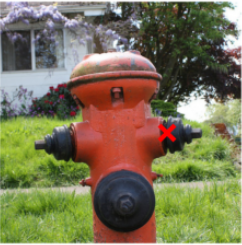} &
        \includegraphics[width=0.235\linewidth,height=3cm,keepaspectratio]{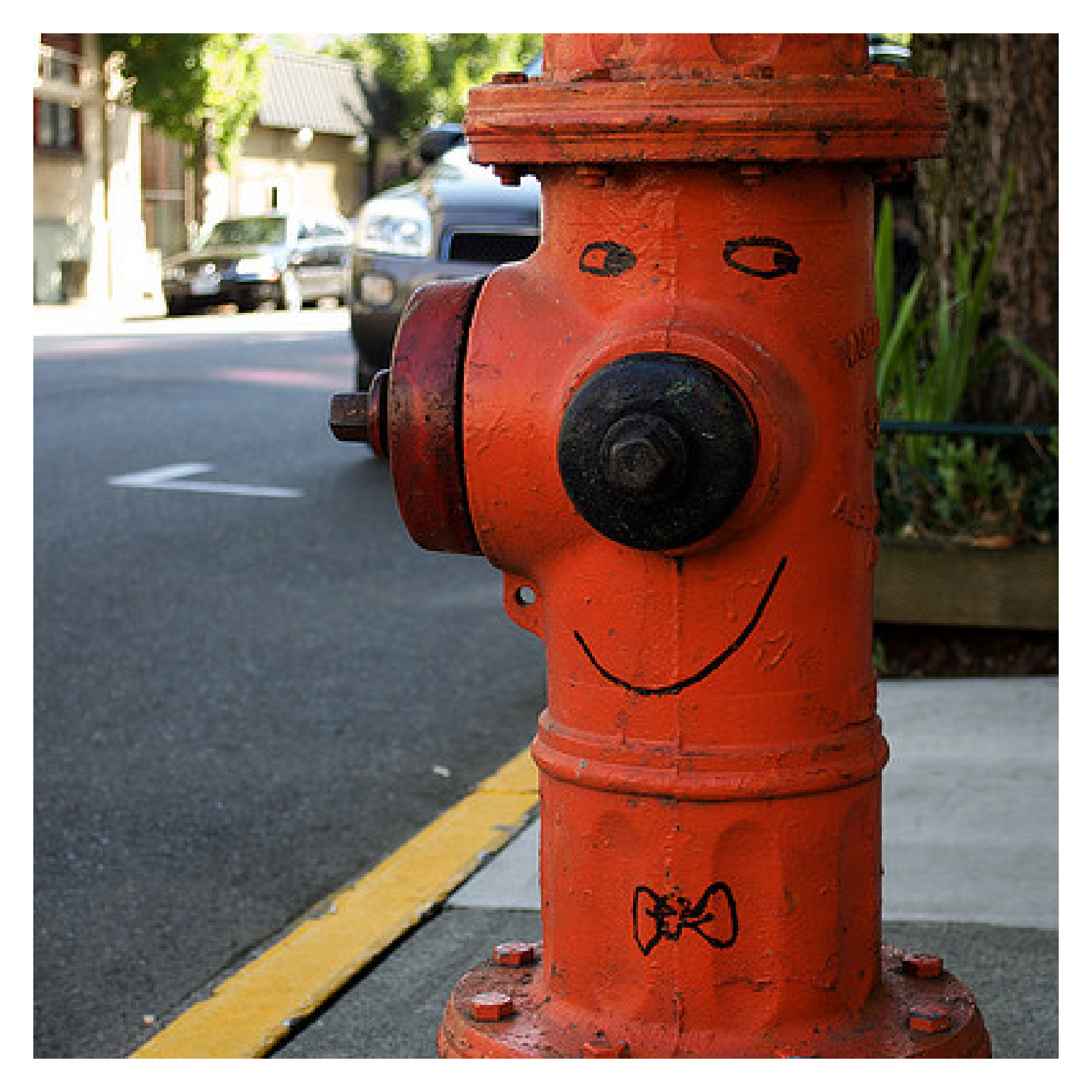} &
        \includegraphics[width=0.235\linewidth,height=3cm,keepaspectratio]{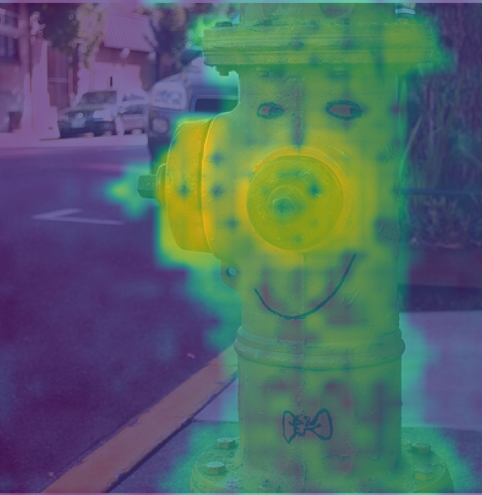} &
        \includegraphics[width=0.235\linewidth,height=3cm,keepaspectratio]{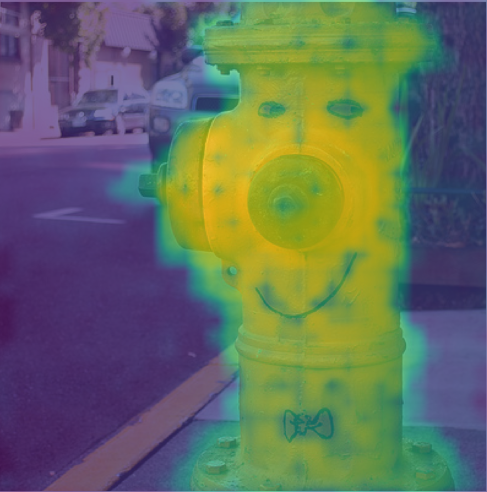} \\
        \includegraphics[width=0.235\linewidth,height=3cm,keepaspectratio]{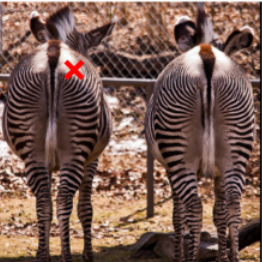} &
        \includegraphics[width=0.235\linewidth,height=3cm,keepaspectratio]{visu/heatmaps/zebra2_resized.png} &
        \includegraphics[width=0.235\linewidth,height=3cm,keepaspectratio]{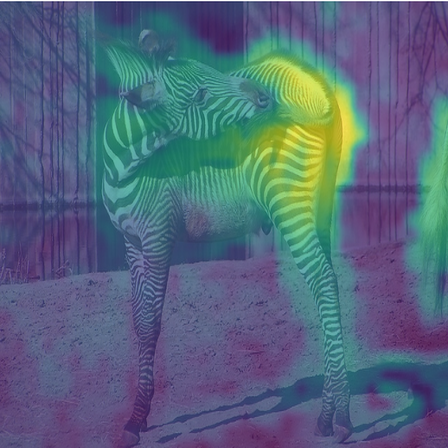} &
        \includegraphics[width=0.235\linewidth,height=3cm,keepaspectratio]{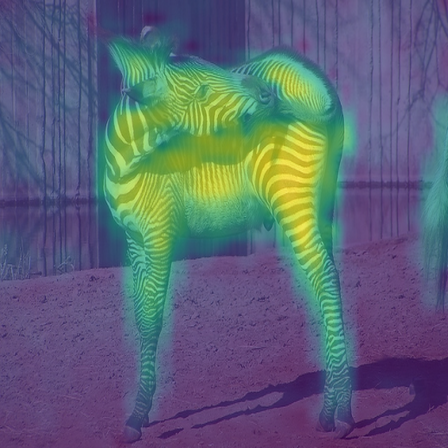} \\
        \includegraphics[width=0.235\linewidth,height=3cm,keepaspectratio]{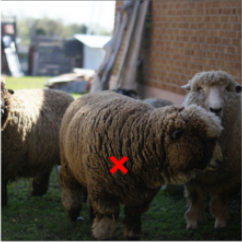} &
        \includegraphics[width=0.235\linewidth,height=3cm,keepaspectratio]{visu/heatmaps/sheep2_resized.png} &
        \includegraphics[width=0.235\linewidth,height=3cm,keepaspectratio]{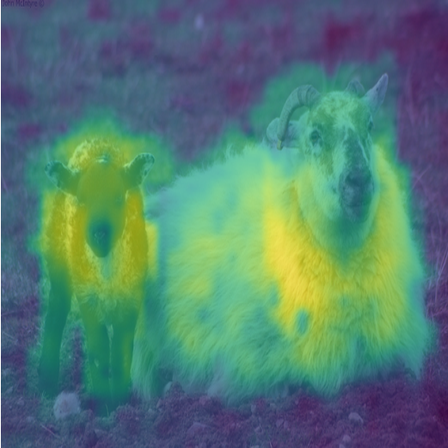} &
        \includegraphics[width=0.235\linewidth,height=3cm,keepaspectratio]{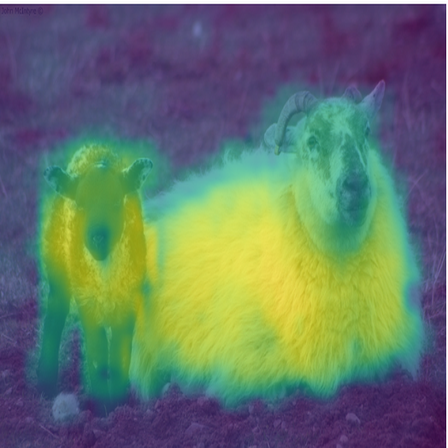} \\
        (a) Query & (b) Reference & (c) DINOv2R & (d) \ours\\
    \end{tabular}
    \vspace{-8pt}
    \caption{Correlation maps between a query image patch (highlighted with a red cross) and a reference image, comparing DINOv2R and \ours. \ours generates coherent, object-level correlations, while DINOv2R produces localized, part-level responses.}
    \label{fig:correlations}
    \vspace{-10pt}
\end{figure}
In \cref{fig:correlations}, we present correlation maps between a query image patch and a reference image, comparing dense representations from DINOv2R and our \ours. While DINOv2R produces localized, part-based correlations, \ours captures semantic-level correspondences, more accurately delineating entire objects of the same semantic type as the query.

This shows that \ours provides better semantic correspondences. 
As visualized in \cref{fig:segmentation_comparison}, this enables \ours to retrieve more semantically coherent nearest neighbors, particularly in low-shot regimes (as in \cref{fig:segmentation_comparison}), resulting in more accurate segmentation outputs compared to DINOv2R.
\section{Conclusion}

We introduced \ours, a novel unsupervised post-training method that enhances dense image representations for in-context scene understanding. Leveraging meta-learning principles and automatically generating in-context tasks with pseudo-segmentations using Stable Diffusion, our approach avoids complex self-distillation architectures. \ours is computationally efficient ($<$9h on an A100) and generalizes well across downstream dense retrieval tasks, including semantic segmentation and depth prediction. It outperforms both the initial pretrained vision encoder and prior state-of-the-art post-training methods, providing a simpler and more effective solution for improving dense representations in vision models.

\clearpage
\paragraph{Acknowledgements}
This work was performed using HPC resources from GENCI–IDRIS (Grants 2025-AD011015037R1, 2024-AD011012884R3, and 2025-AD011016523). 
We thank PEPR Sharp (ANR-23-PEIA-0008, ANR, FRANCE 2030).

{
    \small
    \bibliographystyle{ieeenat_fullname}
    \bibliography{main}
}

\clearpage
\maketitlesupplementary
\appendix

\section{Implementation details}
\label{sec:imp-details}

During post-training, we fine-tune only the last three transformer blocks of the pretrained ViT encoder $f(\cdot)$ while keeping the remaining layers frozen. Our MLP projector $h(\cdot)$ consists of two linear layers with a non-linear activation function GELU~\cite{gelu}. Hidden feature dimension is set to $7D$ and fixed output dimension 6144. $\ell_2$-normalization is applied at the output of the MLP. We set the  temperature parameter $\tau$ of the softmax operator  to $0.07$ (see \cref{tab:ablation_beta}). Our ablation study demonstrates that the method is robust to the choice of temperature, with values ranging from $0.03$ to $0.09$ yielding similar performance on the PascalVOC dataset. During post-training our support set consists of 1 positive and 7 ``distractor'' examples (8 total examples). During training, we use the AdamW~\cite{adamw} optimizer  with a learning rate of  $2.25 \times 10^{-7}$ and a weight decay of $0.05$. We train for $5$ epochs. We employ cosine learning rate schedule.

\begin{table}[h]
 \small
    \centering
   
\begin{tabular}{l c c c}
\toprule
Temperature & 0.09 & \baseline{ 0.07} & 0.03 \\\midrule
PascalVOC & 80.8  & \baseline{81.0} & 81.0\\
\bottomrule
\end{tabular}
    \caption{ Ablation of temperature $\tau$ during post-training.}
\label{tab:ablation_beta}
\end{table}

\paragraph{Generation time of pseudo semantic segmentation labels.}
COCO pseudo-label generation required 17 hours on 1 V100 GPU, with DiffCut inference as the bottleneck due to its current lack of batch processing optimization and suboptimal GPU utilization. However, this one-time offline process can be used to post-train multiple encoders.

\section{Additional quantitative results}

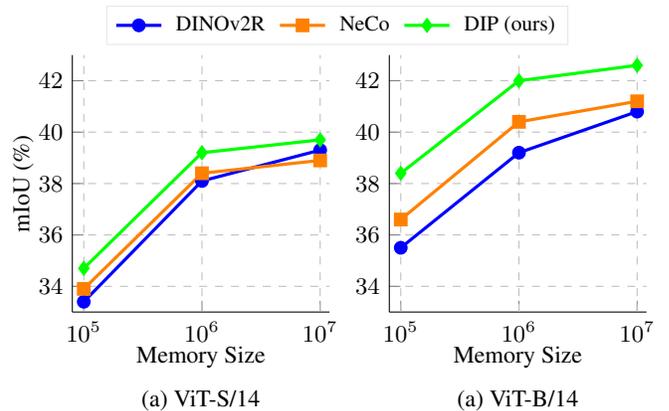
\begin{figure}[t] 
    \centering
    \setlength{\tabcolsep}{0pt} 
    \begin{tabular}{@{}c@{}c@{}}
    \multicolumn{2}{c}{%
        \begin{tikzpicture}
        \centering
        \begin{customlegend}[legend columns=3, legend style={align=center, column sep=.7ex, nodes={scale=0.9, transform shape}, draw=white!90!black}, legend entries={ \small DINOv2R, \small NeCo, \small DIP (ours)}]
            \addlegendimage{color=blue, mark=*, line width=1.2pt}
            \addlegendimage{color=orange, mark=square*, line width=1.2pt}
            \addlegendimage{color=green, mark=diamond*, line width=1.2pt}
        \end{customlegend}
    \end{tikzpicture}
    } \\
    \hspace{-0.05\linewidth}%
        \begin{tikzpicture}
            \tikzstyle{every node}=[font=\small] 
            \begin{axis}[
                axis lines*=left,
                width=0.6\linewidth,
                height=0.6\linewidth,
                xlabel={Memory Size},
                ylabel={mIoU (\%)},
                ylabel style={shift={(0pt,-5pt)}},
                xlabel style={shift={(0pt,5pt)}},
                font=\small,  
                xmode=log,
                log basis x=10,
                xtick={102400,1024000,10240000},
                xticklabels={$10^{5}$,$10^{6}$,$10^{7}$},
                ymajorgrids=true,
                grid style={dashed, gray!50},
                xmin=81920, xmax=12288000,
                ymin=33, ymax=43,
                title={(a) ViT-S/14},
                title style={font=\small},
                title style={yshift=-32.ex,}, 
            ]
                \addplot[
                    color=blue,
                    mark=*,
                    line width=1.2pt,
                ] coordinates {
                    (102400, 33.4)
                    (1024000, 38.1)
                    (10240000, 39.3)
                };
                \addplot[
                    color=orange,
                    mark=square*,
                    line width=1.2pt,
                ] coordinates {
                    (102400, 33.9)
                    (1024000, 38.4)
                    (10240000, 38.9)
                };
                \addplot[
                    color=green,
                    mark=diamond*,
                    line width=1.2pt,
                ] coordinates {
                    (102400, 34.7)
                    (1024000, 39.2)
                    (10240000, 39.7)
                };
                \draw[dashed, gray!50] (axis cs:102400,33) -- (axis cs:102400,43);
                \draw[dashed, gray!50] (axis cs:1024000,33) -- (axis cs:1024000,43);
                \draw[dashed, gray!50] (axis cs:10240000,33) -- (axis cs:10240000,43);
            \end{axis}
        \end{tikzpicture}
    &
        \begin{tikzpicture}
            \begin{axis}[
                axis lines*=left,
                width=0.6\linewidth,
                height=0.6\linewidth,
                xlabel={Memory Size},
                xlabel style={shift={(0pt,5pt)}},
                font=\small,  
                xmode=log,
                log basis x=10,
                xtick={102400,1024000,10240000},
                xticklabels={$10^{5}$,$10^{6}$,$10^{7}$},
                ymajorgrids=true,
                grid style={dashed, gray!50},
                xmin=81920, xmax=12288000,
                ymin=33, ymax=43,
                title={(a) ViT-B/14},
                title style={font=\small},
                title style={yshift=-32.ex,},
            ]
                \addplot[
                    color=blue,
                    mark=*,
                    line width=1.2pt,
                ] coordinates {
                    (102400, 35.5)
                    (1024000, 39.2)
                    (10240000, 40.8)
                };
                \addplot[
                    color=orange,
                    mark=square*,
                    line width=1.2pt,
                ] coordinates {
                    (102400, 36.6)
                    (1024000, 40.4)
                    (10240000, 41.2)
                };
                \addplot[
                    color=green,
                    mark=diamond*,
                    line width=1.2pt,
                ] coordinates {
                    (102400, 38.4)
                    (1024000, 42.0)
                    (10240000, 42.6)
                };
                \draw[dashed, gray!50] (axis cs:102400,33) -- (axis cs:102400,43);
                \draw[dashed, gray!50] (axis cs:1024000,33) -- (axis cs:1024000,43);
                \draw[dashed, gray!50] (axis cs:10240000,33) -- (axis cs:10240000,43);
            \end{axis}
        \end{tikzpicture}
    \hspace*{-0.05\linewidth} 

    \end{tabular}

    \vspace{-10pt}
    \caption{\textbf{In-context scene understanding: impact of memory size.} Semantic segmentation performance on ADE20K (full dataset) using dense nearest neighbor retrieval, evaluated across varying memory sizes.}
    \label{fig:memory_size}
\end{figure}

\parag{In-context scene understanding: impact of memory size.}
In \cref{fig:memory_size}, we analyze the effect of memory size on in-context semantic segmentation using the ADE20K dataset (full) for DINOv2R, NeCo, and our \ours. Results show that \ours outperforms both DINOv2R and NeCo across all memory sizes.
\paragraph{Linear segmentation.}
\begin{table}[t!]
    \small
    \centering
    \begin{tabular}{l l l l}
    \toprule
    Method & Backbone & COCO & ADE20K\\
    \midrule
    DINOv2R & ViT-S/14 &82.1 &33.5 \\
    NeCo & ViT-S/14 & 81.1 (\textcolor{red}{$-$1.0}) & 33.1 (\textcolor{red}{$-$0.4}) \\
    \baseline{\ours (ours)}  & ViT-S/14 & \baseline{\textbf{82.6}  (\textbf{\textcolor{blue}{$+$0.5}})} & \baseline{\textbf{33.7} (\textbf{\textcolor{blue}{$+$0.2})}}\\
    \midrule
    DINOv2R & ViT-B/14 & 85.5 &38.6 \\
    NeCo & ViT-B/14 &85.2 (\textcolor{red}{$-$0.3}) & \textbf{39.5} (\textbf{\textcolor{blue}{$+$0.9}})\\
    \baseline{ \ours (ours) }&  ViT-B/14 & \baseline{\textbf{86.7}  (\textbf{\textcolor{blue}{$+$2.0}})} & \baseline{\textbf{39.5}  (\textbf{\textcolor{blue}{$+$0.9}})} \\
    \bottomrule
    \end{tabular}
    \vspace{-2pt}
    \caption{\textbf{Linear segmentation}
    results on COCO and ADE20K datasets. All methods (DINOv2R, NeCo, and \ours{}) are evaluated using our implementation. \ours{} consistently improves over our base model DINOv2R  and outperforms NeCo across both datasets. }
    \label{tab:lin_segm}
\end{table}
\cref{tab:lin_segm} presents linear segmentation results on COCO and ADE20K benchmarks. For fair comparison, we re-evaluated both DINOv2R and NeCo using our implementation, ensuring consistent evaluation protocols across all methods. Our approach consistently improves over the strong DINOv2R baseline and shows improvements over NeCo. Notably, with the ViT-B/14 backbone on COCO, our method achieves 86.7 mIoU, surpassing DINOv2R by 2.0 points.

\begin{table}[h]
    \centering
\vspace{-7.5pt}

    \begin{tabular}{lccccc} 
        \toprule
      \multirow{2}{*}{\textbf{Method}} & \multirow{2}{*}{\textbf{Backbone}}& \multicolumn{4}{c}{\textbf{PascalVOC}} \\
     & &1  &  $\frac{1}{8}$ & $\frac{1}{64}$ & $\frac{1}{128}$  \\
        \midrule
        DINOv2R& ViT-B & 79.0 & 75.3 & 67.6 & 60.3 \\
         \rowcolor{baselinecolor} \ours (ours) & ViT-B& 82.1 & \textbf{79.6} & \textbf{75.1} & \textbf{70.1} \\
        \midrule
        DINOv2R  & ViT-L &76.9 &72.8 & 61.4& 54.4 \\
        \rowcolor{baselinecolor} \ours (ours) & ViT-L& \textbf{81.1}&\textbf{78.7}  & \textbf{70.0} & \textbf{64.6} \\
      
        \bottomrule
    \end{tabular}
    \caption{\textbf{Larger backbones evaluation} We show performance of \ours and DINOv2R on a larger backbone ViT-L. Dense nearest neighbor retrieval performance (mIoU) for semantic segmentation on PascalVOC across varying proportions of training data.}
    \label{tab:vitl}
\end{table}

\paragraph{Larger backbones.}
While ViT-L (DINOv2) underperforms ViT-B in in-context segmentation \cite{pariza2024hbird, hummingbird}, 
our method still improves results with ViT-L, as shown in \cref{tab:vitl}. This demonstrates \ours's scalability across backbone sizes.

\begin{table}[h]
    \centering
    \scriptsize

    \begin{tabular}{lcccccccc} 
        \toprule
     &  \multicolumn{4}{c}{\textbf{PascalVOC}} & \multicolumn{4}{c}{\textbf{ADE20K}}\\
        &  $1$ & $\frac{1}{8}$ & $\frac{1}{64}$ & $\frac{1}{128}$   & $1$ & $\frac{1}{8}$ & $\frac{1}{64}$ & $\frac{1}{128}$  \\
        \midrule
        Two Crops    &79.5 & 75.1& 67.7 & 61.2&39.4&33.2&24.7&22.4\\
        \rowcolor{baselinecolor}\cellcolor{white} NN & \textbf{81.0}  & \textbf{77.7} & \textbf{71.4} &\textbf{65.9}&\textbf{39.7}&\textbf{33.7}&\textbf{25.6}&\textbf{23.2}\\
      
        \bottomrule
    \end{tabular}
        \caption{\textbf{Additional ablation on the construction of positive examples}. Dense nearest neighbor retrieval performance (mIoU) for semantic segmentation on PascalVOC and ADE20K across varying proportions of training data.}

    \label{tab:nn_vs_twocrops}
    \end{table}

\paragraph{Nearest Neighbors (NN) vs. Two Crops}
We compare two strategies for creating positive examples: (1) retrieving nearest neighbors using DINOv2R image-wise features (NN) and (2) using two random crops from the same image.
NN consistently outperforms Two Crops, with the performance gap widening when fewer training examples are available (see \cref{tab:nn_vs_twocrops}). This scalability advantage justifies our design choice.

\section{Additional qualitative results}
\begin{figure*}[t!]
  \centering
  \includegraphics[width=\textwidth]{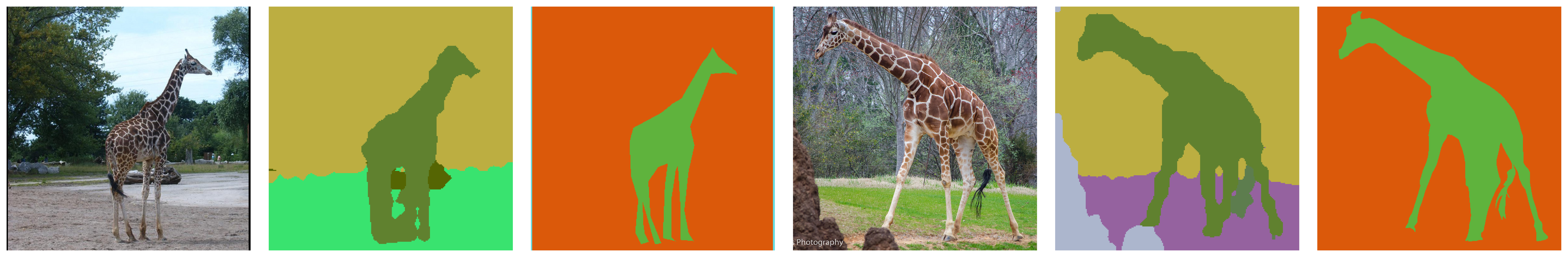}\\[1ex]
  \includegraphics[width=\textwidth]{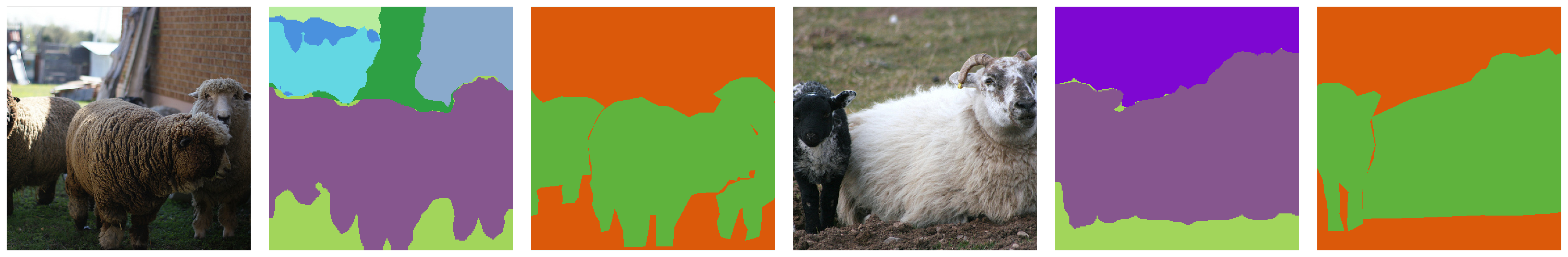}

  \begin{minipage}{\textwidth}
    \centering
    \begin{minipage}{0.15\textwidth}
      \centering
      \subcaption{Query image}
    \end{minipage}\hfill
    \begin{minipage}{0.15\textwidth}
      \centering
      \subcaption{Pseudo labels}
    \end{minipage}\hfill
    \begin{minipage}{0.15\textwidth}
      \centering
      \subcaption{GT labels}
    \end{minipage}\hfill
    \begin{minipage}{0.15\textwidth}
      \centering
      \subcaption{Positive image}
    \end{minipage}\hfill
    \begin{minipage}{0.15\textwidth}
      \centering
      \subcaption{Pseudo labels}
    \end{minipage}\hfill
    \begin{minipage}{0.15\textwidth}
      \centering
     \subcaption{GT labels}
    \end{minipage}
  \end{minipage}
 \caption{\textbf{Examples of automatically constructed in-context scene understanding tasks.} Each row shows a query image and its corresponding positive support example. (a) and (b) display the query image and its pseudo segmentation labels, while (d) and (e) show the positive support image and its pseudo segmentation labels. (c) and (f) present the ground truth segmentation labels for the query and positive images, respectively, included only for comparison with the pseudo labels. }
  \label{fig:pseudolabel_labels_suppmat}
\end{figure*}

We present additional examples of automatically constructed in-context tasks in \cref{fig:pseudolabel_labels_suppmat}, showing the quality of our pseudo-labeling approach. We display query images paired with their corresponding positive support examples, along with both pseudo-labels and ground truth labels, included only for comparison. 


\end{document}